\documentclass[letterpaper, 10 pt, journal, twoside]{IEEEtran}
\usepackage{amssymb, amsthm, amsmath,  cite, color}
\usepackage{algorithmic}
\usepackage{enumerate}
\usepackage{booktabs}
\usepackage{footmisc}
\usepackage{balance}
\usepackage{bm}
\usepackage[dvipsnames]{xcolor}
\usepackage{float}
\usepackage[font={small}]{caption}
\usepackage{graphicx}
\usepackage{hyperref}
\usepackage{import}
\usepackage{mathtools}
\usepackage{multirow}
\usepackage{outline}
\usepackage{paralist}
\usepackage{siunitx}
\usepackage{stmaryrd}
\usepackage{systeme}
\usepackage{subcaption}
\usepackage{tikz}
\usepackage{titlesec}
\usepackage{hyperref}
\usepackage{tabularx}

\usepackage{url}
\usepackage{threeparttable}

\usepackage[linesnumbered,ruled,vlined]{algorithm2e}
\usepackage{algorithm2e}
\usepackage{algorithmic}

\usepackage[font={small}]{caption}
\SetAlFnt{\small}
\SetAlCapFnt{\small}
\SetAlCapNameFnt{\small}
\algsetup{linenosize=\small}
\newtheorem{remark}{Remark}
\newcommand{\revised}[1]{{\color{black} #1}}

\title{\LARGE \bf
Fast Path Planning for Autonomous Vehicle Parking with Safety-Guarantee using Hamilton-Jacobi Reachability}

\author{
\IEEEauthorblockN{Xuemin Chi$^{1}$, Jun Zeng$^{2}$, Jihao Huang$^{1}$, Zhitao Liu$^{1*}$, Hongye Su$^{1}$}
\thanks{This work was supported in part by National Key R\&D Program of China (Grant NO. 2021YFB3301000), National Natural Science Foundation of China (NSFC:62173297) and Zhejiang Key R\&D Program (Grant NO. 2022C01035).

Copyright (c) 2026 IEEE. Personal use of this material is permitted. However, permission to use this material for any other purposes must be obtained from the IEEE by sending a request to pubs-permissions@ieee.org.

\IEEEauthorblockA{$^1$ State Key Laboratory of Industrial Control Technology, Institute of Cyber-Systems and Control, Zhejiang University, Hangzhou, China {\tt\small  \{chixuemin, jihaoh, ztliu, hysu\}@zju.edu.cn}.} 

\IEEEauthorblockA{$^2$ Hybrid Robotics Group at the Department of Mechanical Engineering, UC Berkeley, USA {\tt\small  \{zengjunsjtu\}@berkeley.edu}.}

\IEEEauthorblockA{$^*$ Corresponding author.}

}
}

\begin{document}

\newtheorem{Proposition}{Proposition}
\renewcommand\qedsymbol{$\blacksquare$}

\maketitle

\thispagestyle{empty}

\begin{abstract}
We present a fast path planning algorithm called Hamilton-Jacobi-based bidirectional A* (HJBA*) for solving general tight parking scenarios. The algorithm consists of two layers: a high-level Hamilton-Jacobi (HJ) reachability analysis and a lower-level bidirectional A* search. In the high-level reachability analysis, a backward reachable tube, considering vehicle dynamics, is computed using HJ analysis. This tube intersects with a safe set to define a safe reachable set. The safe set is determined by constraints of positive signed distances from obstacles in the environment, which are computed offline by solving quadratic programming problems.
For states within the safe reachable set, the backward reachable tube ensures that they are reachable under system dynamics and input bounds, while the safe set guarantees safety with respect to obstacles of varying shapes. For online computation, randomized states are sampled from the safe reachable set and used as heuristic guide points in the bidirectional A* search. The bidirectional A* search is parallelized for each randomized state from the safe reachable set.
We demonstrate that the proposed two-level planning algorithm effectively and efficiently solves various parking scenarios for typical parking requests. Simulations in large-scale randomized parking scenarios validate the algorithm’s performance, showing that it outperforms other state-of-the-art parking planning algorithms.
The real-world experiment is conducted to further validate the effectiveness of our approach.
\end{abstract}

\begin{IEEEkeywords}
Hamilton-Jacobi reachability, path planning, autonomous parking, planning in constrained environments
\end{IEEEkeywords}

\section{Introduction}
\label{sec:intro}

\subsection{Motivation}
\IEEEPARstart{A}{utonomous} parking has long been a crucial topic for self-driving vehicles and
Intelligent Parking Assist System (IPAS) has become an increasingly crucial component of Advanced Driver Assistance Systems (ADAS)~\cite{gruyer2017perception}.

\begin{figure}
\centering
\includegraphics[width=1\linewidth]{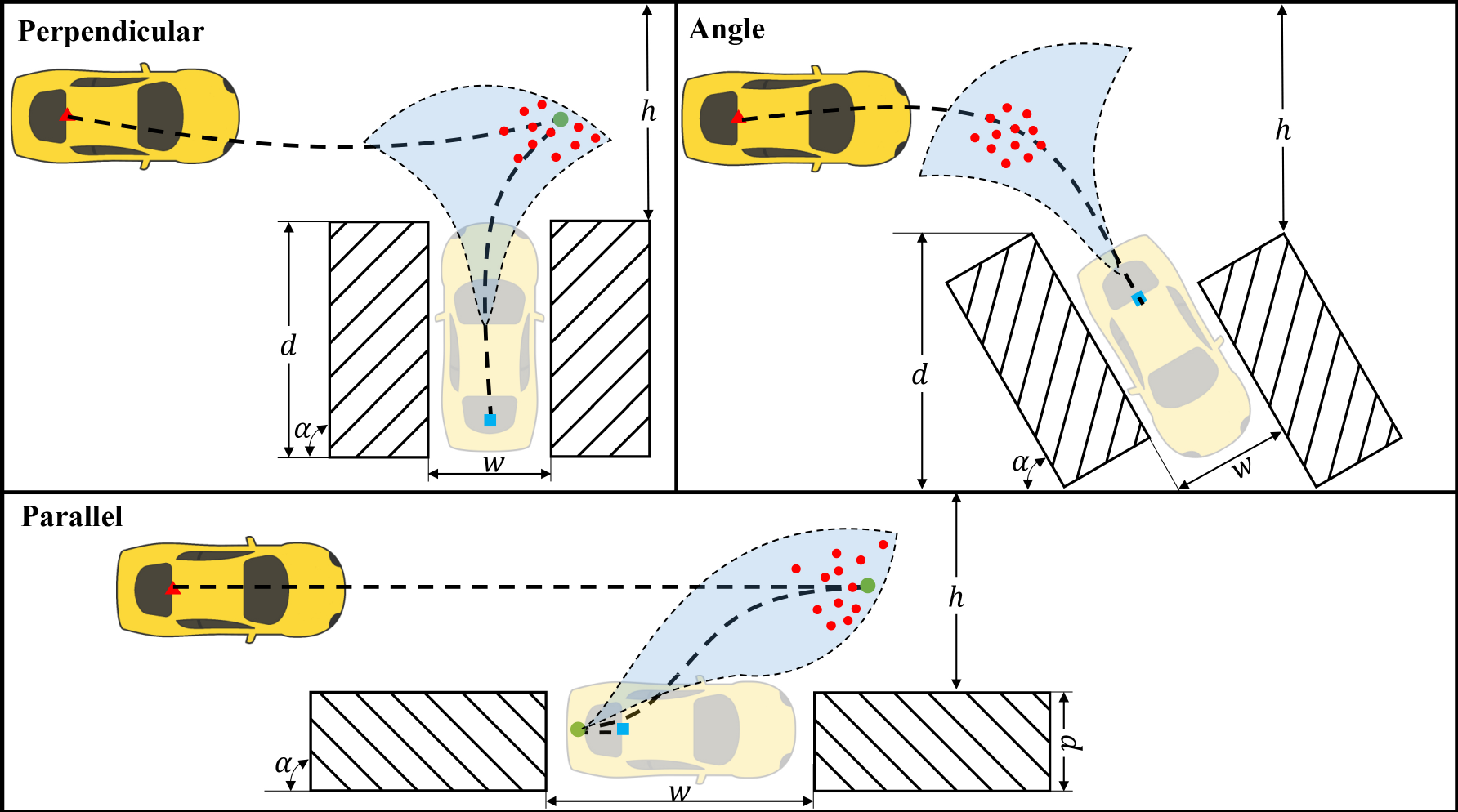}
\caption{Three typical types of common parking scenarios: Perpendicular, Angle, and Parallel. 
The initial pose is marked in a red triangle and the goal pose is marked in a blue square.
The solid green dot is the cusp where a direction changes.
The light blue domain is a subset of the BRT, which is used to guide the search.
The red circles are random connected states in the BRT.
$h$ is the height of free driving space in a parking lot, $d$ is the distance between the inner side of the parking spot and the outer side towards the free driving area, and $w$ is the width of a parking spot.}
\label{fig:demonstration}
\end{figure}
Autonomous parking in a dense environment is challenging since the computational complexity of optimization-based~\cite{zhang2020optimization} planning algorithm is dramatic due to clustered obstacles.
In order to economize it, existing work usually exploits a global planning algorithm where search-based planning is used to find a kinetically feasible path between the initial state and the target state, such as Hybrid-A*~\cite{dolgov2008hybridA}.
However, these existing approaches could suffer from deadlock~\cite{thirugnanam2022duality} or unexpected timeout due to the heuristic cost function in search-based planning algorithm design.
Therefore, these approaches can not be applied to solve an online planning problem.

In practice, the parking environment itself is usually fixed, including surrounding obstacles and a finite number of potential parking spots, and we expect to handle various parking requests quickly under the same parking environment with different initial states.
Each target state is associated with each parking spot, including vehicle position and orientation.
In this paper, we are motivated by this fixed parking environment setup to propose a novel planning architecture, where parking path planning problems are decoupled into offline reachability calculation and online search calculation.
Our offline reachability analysis considers finite potential target states, where the safe reachable set of each target state is calculated.
The safe reachable set is the intersection of a backward reachable tube (BRT), calculated from Hamilton-Jacobi (HJ) reachability analysis with respect to the vehicle state, and the safe set is defined by signed distance constraints according to the vehicle's rectangular geometry.
Then, given a target state, multiple connected states are sampled from this safe reachable set, and a bidirectional heuristic search is used to find kinematically feasible two-segment paths between the initial state to a connected state, and a connected state to the target state, respectively.
An optimal path could be selected from these generated two-segment paths depending on the user's preferences.
These sampling connected states allow us to avoid excessive search compared with existing work.
Our offline/online planning architecture allows us to solve various parking scenarios without any timeout, shown in Fig.~\ref{fig:demonstration}, and with lower computational complexity than the state-of-the-art.

\subsection{Related Work}

 Motion planning is a very active research area in robotics communities and autonomous parking planning is a subdomain for non-holonomic robots~\cite{wang2014automatic}.
Generally, planning algorithms fall into five categories: graph-based~\cite{hart1968astar,likhachev2005anytime,nash2007theta}, sampling-based~\cite{lavalle1998rapidly,cortes2021sampling,lavalle2001randomized}, curves interpolating~\cite{he2022race}, trajectory optimization~\cite{zhang2020optimization,zips2013fast,leu2022autonomous,chi2022optimization}, and reachability analysis~\cite{mitchell2005time,tomlin1996hybrid,chen2016multi,fisac2015reach,bansal2017hamilton}.

Sampling-based approaches (e.g., RRT*~\cite{lavalle2001randomized}) show an excellent scale performance in high dimensional path planning. 
Nevertheless, classic sampling-based approaches cannot plan a combination of forward and backward maneuvers.
Therefore, in~\cite{banzaf2017hybrid,jhang2020forward}, a bi-directional RRT-based approach is extended with Reeds-Sheep (RS) curves~\cite{reeds1990optimal} and applied in autonomous parking. 
In recent work~\cite{dong2020knowledge}, the Gaussian Mixture Model (GMM) is utilized to learn the distribution of sampling seeds, which subsequently guides the sampling process. However, this framework is limited to short-horizon planning problems and suffers from non-competitive overall runtime performance.
Curves (e.g., Bezier curves~\cite{he2022race}) interpolating approaches are preferred because of smoothness and low computation through picking an optimal candidate curve, which further enhances their applicability in real-time planning scenarios.

Optimization-based methods such as TrajOpt~\cite{schulman2014ootion}, CHOMP~\cite{zucker2013chomp}, LQR~\cite{lembono2021probabilistic} and MPC~\cite{dantec2021whole} based methods (aka. Trajectory optimization) formulate the motion planning problem as an optimal control problem that contains the objective function and various constraints. 
The problem can be solved directly by gradient-based iterative methods or reformulated as a numerical optimization problem solved by general solvers~\cite{kondak2001computation} (e.g., Gurobi, Cplex, IPOPT). 
Early work~\cite{li2015unified} formulated the parking problem as an optimization and solved it by the IPOPT solver.
While optimization-based approaches own the advantages that they can yield very smooth and good results sometimes and handle user-defined nonlinear constraints, the quality of the results is strongly related and sensitive to the initial guess~\cite{lembono2020memory}. 
In work~\cite{qiu2021hierarchical}, the authors employed the Gauss pseudo-spectral method to solve an optimization problem, generating a parking reference path to the MPC controller.
Work~\cite{li2016time} also adopts the strategy of using an optimization-based method to provide the initial guess to better solve the parking problem.
An offline dataset is built by sequentially solving an optimization problem for online receding horizon control.
In \cite{zips2016optimisation}, rather than providing an initial guess, the author carries out an optimization strategy when the car is near the parking spot to make the optimization problem solvable.
Graph-based methods can also be used to provide an initial guess for efficient optimization~\cite{wang2024hierarchical}.

Another branch of methods is graph-based and geometric-based approaches.
graph-based search methods are widely used in robot navigation (e.g, A*~\cite{hart1968astar} and its variants~\cite{likhachev2005anytime}), and predefined motion primitives allow considering the kinematic constraints when searching.
Early pure geometric works rely on Dubins curves \cite{dubins1957curves} and RS curves to plan a parking path but they are limited to obstacle-free environments.
On the other hand, instead set of straight lines and circular arcs, some work~\cite{cai2022geometric,vorobieva2013geometric,sungwoo2011geometric} focus on planning a continuous-curvature parking path and different curves are used, nevertheless, these approaches suffer from sub-optimal parking length, fixed or limited start configuration, and they are also limited to obstacle-free scenarios. 
Most A*-based variants such as ARA*~\cite{likhachev2005anytime}, and Theta*~\cite{nash2007theta} are unable to handle a parking problem for autonomous vehicles in practical scenarios. 

One classic and prevailing approach is Hybrid A* (HA*)~\cite{dolgov2008hybridA}, which predefined a 3D ($x, y, \theta$) node motion primitives subjected to vehicle kinematic constraints and incorporates the RS curves. 
However, the complex heuristic function requires a carefully manual trade-off between different penalty terms like backward movements, directional switches, turning movements, and others.
In recent works \cite{zhang2020optimization,zhou2020autonomous, han2023efficient}, HA* is used to speed up trajectory optimization in parking problems and guarantee feasibility.
The limitation is that due to the low update frequency of HA* (e.g., worse cases from seconds to minutes), the performance of the whole pipeline is inefficient.
Work \cite{abbeel2008apprenticeship} shows that the trade-off from cost function (e.g., forward, backward, and turning search costs) in HA* also leads to an unstable performance in different parking scenarios. 
HA* has been used to generate path candidates in~\cite{zhang2022robust}, 
the issue is when the start and target change, the inefficient problem brought by HA* still exists so that the update frequency is very low.
In~\cite{sedighi2019guided}, the authors use a visibility graph to find the shortest path at first and then use it to guide the HA*.
The disadvantage is that it can be time-consuming to run the visibility graph algorithm and the conservative path is often useless in tight parking environments.
We exploit the reachable set which can find a solution in the tight parking environments if there exists one.
The multi-heuristic hybrid A* (MHHA*) ~\cite{huang2022search} has been proposed which is an extension based on~\cite{aine2016multi}.
The limitation of MHHA* is that it requires more parameter tuning and the balance between different heuristic functions.
Our algorithm exploits a new node expansion strategy to cooperate with the reachable set to avoid heavy parameter tuning.

With the observation that the reachability analysis technique enables the computation of a global optimal backward reachable tube that can serve as heuristic guidance for search-based algorithms. 
This allows for the repeated use of the offline computed BRT to enhance search speed and ensure safety, thereby offering a solution to the challenges outlined above. 
Hamilton-Jacobi (HJ) reachability analysis, a verification method for the safety and performance of a system, has recently become popular in motion planning and collision avoidance applications~\cite{bansal2017hamilton,fisac2015reach,chen2016multi}. In HJ reachability analysis, the system is represented by an ordinary differential equation (ODE) and a target set describes the set of unsafe states or desired states~\cite{tomlin1996hybrid}.
The BRS/BRT can be acquired by solving HJ partial differential equations (PDEs) and accurate numerical methods in level set methods~\cite{fedkiw2002level} provided fast solution schemes for HJ PDEs.
In work~\cite{chen2021fastrack}, a framework for planning and tracking on-the-fly is proposed and the HJ PDEs are computed to acquire the TEB of a relative system between planning and tracking.

\subsection{Contributions}
Prior work on path planning for autonomous parking has not been able to provide a real-time path planning algorithm for complex environments. 
To address this challenge, we propose a novel and efficient two-layer planning strategy that builds on Hamilton-Jacobi reachability analysis and a graph-based search algorithm. 
Our work makes several key contributions:

\begin{itemize}
    \item We present a path planning algorithm that leverages HJ reachability analysis with a graph-based search approach to provide fast planning in general parking scenarios. 
    
    \item Our path planning approach is designed to provide fast and stable planning in various parking scenarios e.g., multiple parking requests without parameter tuning.
    
    \item We validate the performance of our proposed algorithm through real-world experiments and benchmark it against state-of-the-art planning algorithms. 
    The results show that our proposed algorithm outperforms other algorithms in terms of efficiency and stable performance.
\end{itemize}

\subsection{Paper Structure}
The structure of this paper is as follows. We begin by presenting the framework of our algorithm HJBA* in Section~\ref{sec: framework}. Next, we provide a detailed explanation of the process of computing the safe reachable set in Section~\ref{sec: offline}. Section~\ref{sec: online} describes the process of bidirectional search to the connected states. In Section~\ref{sec: results}, we present the results of our big-scale batch simulations in various parking scenarios and benchmark them against other state-of-the-art planning algorithms. Finally, we conclude the paper with remarks on potential future research directions and open problems in Section~\ref{sec: conclusion}.

\begin{table}[!h]
    \centering
        \caption{Variable Table.}
    \resizebox{\columnwidth}{!}
    {%
    \begin{tabular}{l l}
    \hline
    Notations & Descriptions   \\ \hline
    $\mathbf{z}$ & vehicle states vector contains $x, y, \theta$\\
    $\Omega_{\mathbf{z}}$ & discretized domain space along each dimension of $\mathbf{z}$\\
    $\mathbf{z}_{0}, \mathbf{z}_{g}$ &  initial and goal state vector of vehicle system   \\
    $\mathcal{G}_{0}$ &  parking goal set \\ 
    ${\mathcal{G}}$ & backward reachable tube \\
    $\mathcal{C}$ &  safe signed distance set \\ 
    $\mathcal{S}$ &  safe reachable set\\
    $R(\cdot), \mathbf{T}(\cdot)$ &  rotation matrix and translation vector \\ 
    $A_{i}$, $b_{i}$ & $A_{i} y_i \leqslant b_i$, describing a polytopic area  \\
    $\mathbb{V}$ & the geometric space occupied by a car described by $A_i$ and $b_i$ \\
    $\mathbb{O}_{i}$ & $i$-th obstacle described by $A_i$ and $b_i$\\
    $g(z_{0}, z_{\text{g}})$ &  computed cost-to-go from $z_{0}$ to $z_{\text{g}}$   \\ 
    $h(z_{0}, z_{\text{g}})$ &  estimated cost-to-come from $z_{0}$ to $z_{\text{g}}$   \\ 
    $c(\cdot)$ &  cost of a node composed of $g(\cdot,\cdot)$ and $h(\cdot,\cdot)$ \\
    $\mathcal{A}(z_{0}, z_{\text{g}})$ &  path connected $z_{0}$ and $z_{\text{g}}$ generated by node expansions \\ 
    $\mathcal{RS}(z_{0}, z_{\text{g}})$ &  path connected $z_{0}$ and $z_{\text{g}}$ generated by RS curves \\ 
    $P(\mathbf{z}_{0}, \mathbf{z}_{g})$ &  path starts at $z_{0}$ and ends $z_{\text{g}}$ \\ 
    \revised{$Q_{\textup{open}},Q_{\textup{close}}$ }& \revised{Open and close list maintained in forward or backward searching }\\
    \revised{$\mathbf{z}_{k}^{\mathcal{S}}$} & \revised{Sampling connected states from $\mathcal{S}$}\\
    \revised{$\mathbf{z}_{\textup{best}}$} & \revised{States connected to the goal with collision-free $\mathcal{RS}$ curves}\\
    \revised{$\mathbf{z}^{\textup{F}}, \mathbf{z}^{\textup{B}}$ } & \revised{States expanded in forward search and backward search} \\\hline
    \end{tabular}
    }
    \label{tab:variable_list}
\end{table}

\section{Framework of HJBA*}
\label{sec: framework}
\begin{figure*}[!t]
\centering
\includegraphics[width=1\linewidth]{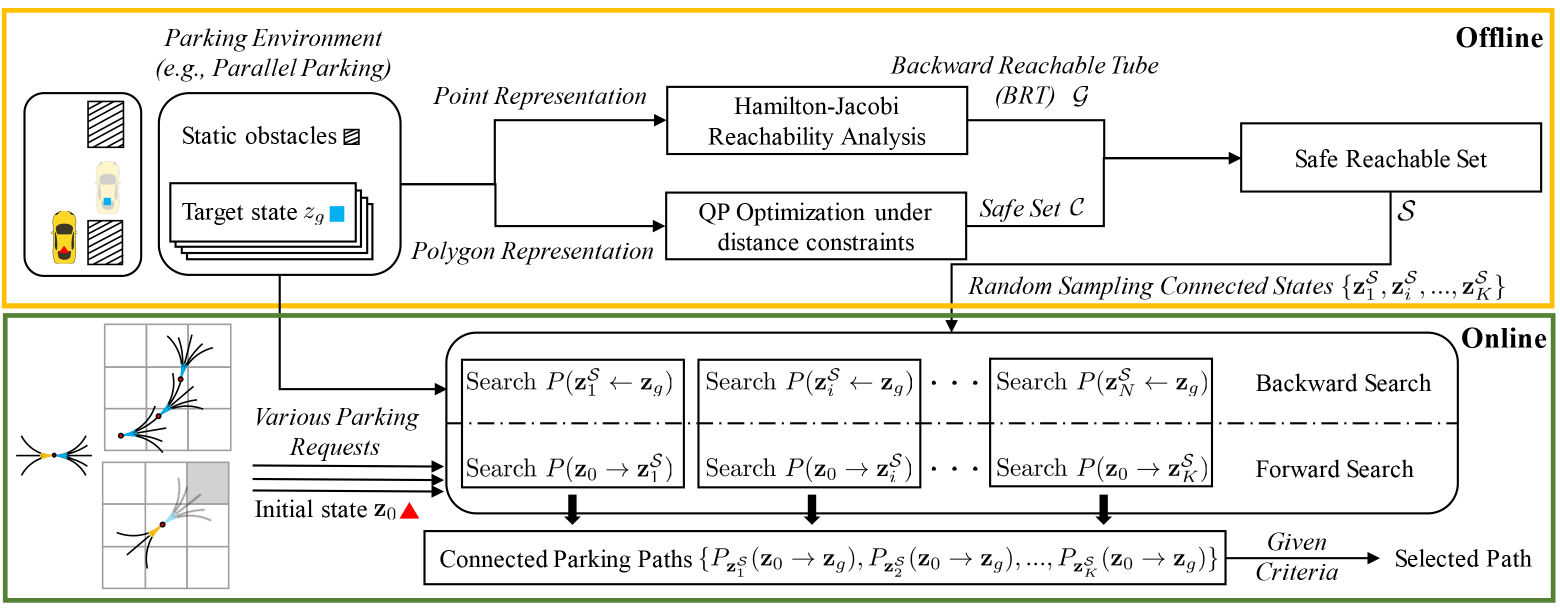}
 \caption{Our approach involves a two-layer scheme. The parking environment includes the parking spot and obstacles involved in information.
  In offline computation, we compute the safe reachable set to provide sampling connected states.
  In online searching layer, we perform a parallel bidirectional search to handle various parking requests.
}
\label{fig:general_framework}
\end{figure*}

This section presents the workflow of HJBA* and summarizes the details in Fig.~\ref{fig:general_framework}. Our algorithm aims to efficiently find an optimal parking path for a car in a parking lot. 
In the event that the desired parking spot is already occupied, a fast replan strategy is employed in our bidirectional search-based algorithm to quickly find an alternative parking spot. 
Furthermore, during the parking maneuver, drivers usually approach the parking spot gradually, making constant adjustments to their position while prioritizing safety before successfully parking the car. 
To incorporate this heuristic and safety-aware strategy into our algorithm, we introduce the concept of a safe reachable set and implement it in our offline layer.

The offline component of our algorithm aims to generate a safe reachable set, denoted by $\mathcal{S}$, that can be utilized by the online framework. 
This set comprises states that satisfy both safety and feasibility requirements. Specifically, $\mathcal{S}$ is obtained by taking the intersection of the Backward reachable tube, denoted by $\mathcal{G}$, and the safe set, denoted by $\mathcal{C}$. 
To perform reachability analysis in our algorithm, we adopt the Dubins car model due to its simplicity and computational efficiency, which does not require considering the vehicle's dynamics and frictions at the global planner level. 
However, this approach utilizes a point representation that fails to account for the geometric shape of the car. 
As a result, states in $\mathcal{G}$, are subject to system dynamics but may not guarantee safety in tight environments. 
To overcome this limitation, we address collision avoidance between the car's convex rectangular shape and the environment by solving a quadratic programming (QP) optimization problem. 
This optimization problem incorporates the constraint of positive signed distance to ensure safety.

In practice, the target state, denoted by $\mathbf{z}_g$ in our algorithm refers to one of the finite parking spots available in the parking environment. The information about the parking environment is utilized to generate both $\mathcal{G}$ and $\mathcal{C}$. 
The intersection of these sets forms $\mathcal{S}$, which is then utilized by the online component of our algorithm. 
To create heuristic connected states for the online framework, we randomly sample states from $\mathcal{S}$. 

During the online phase, our algorithm allows any initial state $\mathbf{z}_0$ within the parking environment to be specified based on the users' parking requests. Utilizing the information about $\mathbf{z}_0$ and the target state $\mathbf{z}_g$, we perform a parallel bidirectional A* search to obtain the connected states, denoted by $\mathbf{z}^{\mathcal{S}}_i$. The bidirectional A* search involves both forward and backward searches. In the forward search, we compute the paths that connect $\mathbf{z}_0$ and $\mathbf{z}^{\mathcal{S}}_i$, denoted by $P(\mathbf{z}_0 \rightarrow \mathbf{z}^\mathcal{S}_i)$. In the backward search, we compute the paths that connect $\mathbf{z}_g$ and $\mathbf{z}^{\mathcal{S}}_i$, denoted by $P(\mathbf{z}^\mathcal{S}_i \leftarrow \mathbf{z}_g)$. By combining each pair of these paths, we obtain the connected parking paths denoted by $P_{\mathbf{z}_{i}^{\mathcal{S}}}(\mathbf{z}_0 \rightarrow \mathbf{z}_g)$. Based on the criteria specified by the users, one of these connected paths is selected and output as the optimal path.

\section{Safe reachable set computation}\label{sec: offline}
This section introduces offline computation of HJBA*. 

\subsection{Backward Reachable Tube}
Consider a continuous-time control system, which evolves according to the ODE as follows:
\begin{equation} \label{eq:system_dynamics}
\begin{array}{cc}
     \dfrac{d \mathbf{z}(\tau)}{d \tau}=\dot{\mathbf{z}}(\tau)=f(\mathbf{z}(\tau), u(\tau)), \tau\in [t,T]  \\
     \mathbf{z}(\tau)\in \mathcal{Z}, u(\tau) \in \mathcal{U},
\end{array}
\end{equation}
where $\mathbf{z}(\tau) \in \mathcal{Z}\subseteq \mathbb{R}^{n}$ represent the state of the system model. The control input is denoted by $u(\tau)$ and $\mathcal{U} \subseteq \mathbb{R}^{m}$. 
The control function $u(\cdot)$ is drawn from the set of measurable functions: $(t,T) \rightarrow \mathbb{U}$.
where $\mathbb{U}$ is compact and satisfies the safety control constraints. 
The system dynamics $f: \mathbb{R}^{n} \times \mathcal{U} \rightarrow \mathbb{R}^{n}$ is assumed to be Lipschitz continuous uniformly in $\mathbf{z}$ for fixed $u$. 
Therefore, Given a $u(\cdot) \in \mathbb{U}$, there is always a unique trajectory starting from state $\mathbf{z}$ at time $t$ denoted as $\mathcal{L}_{\mathbf{z},t}^{u}(\tau): [t,T] \rightarrow \mathbb{R}^{n}$ under these assumptions satisfying~\eqref{eq:system_dynamics}. 

A BRT represents the set of states $z\in \mathbb{R}^{n}$ from which the system can be driven into a goal set $\mathcal{G}_{0}$ within a time horizon $\left| T-t \right|$. 
It is formally described as:
\begin{equation}
\mathcal{G}(t)=\{\mathbf{z}: \forall u(\cdot) \in \mathbb{U}, \exists \tau \in [t,T],
\left.\mathcal{L}_{\mathbf{z},t}^{u}(\tau) \in \mathcal{G}_0\right\},
\end{equation}
We often represent our goal set $\mathcal{G}_0$ by a sublevel set function $l(\mathbf{z}): \mathcal{Z} \rightarrow \mathbb{R}$ as $\left\{\mathcal{G}_{0} ={\mathbf{z}\in \mathcal{Z} : l(\mathbf{z}) \leq 0}\right\}$.

\begin{remark}
In the context of HJ reachability, a backward reachable set can represent either a target or an unsafe set. 
In this paper, we consider it to represent a target set and utilize it to model finite parking spaces in a given parking lot, as depicted in Fig.~\ref{fig:general_framework}
\end{remark}

For the details on how to formulate the Hamilton-Jacobi-Isaacs Variation Inequality and the level-set solution scheme used in our algorithm, please refer to~\cite{fisac2015reach}.
The system model used is
\begin{equation} \label{eq:dubins}
    \dot{\mathbf{z}}=
    \left[\begin{array}{c}
    \dot{x} \\
    \dot{y} \\
    \dot{\theta}
    \end{array}\right]=\left[\begin{array}{c}
    v \cos \theta \\
    v \sin \theta \\
    \omega
    \end{array}\right],
\end{equation}
where $(x,y,\theta)$ represents the position and heading angle in Cartesian coordinates, $(v, w)$ is the constant speed and control input angular speed.

\subsection{Safe Set}
Consider the $\mathcal{G}$ is computed over a fixed Cartesian grid in $\mathbb{R}^3 \times [T,t]$.
The grid is denoted by $\Omega_{\mathbf{z}} = \Omega_x \times \Omega_y \times \Omega_{\theta}$.
 For simplicity, assume we discretize each bounded interval by $n$ number of elements.
For a state $\mathbf{z}_k \in \Omega_z$, we denote the geometry space occupied by:
\begin{equation}
    \mathbb{V}_k :=  R(\mathbf{z}_{k})\mathbb{V}_{0} + \mathbf{T}(\mathbf{z}_{k}),
\end{equation}
where $R(\cdot) \in SE(2) $ is a rotation matrix, and $\mathbf{T}(\cdot): \mathbb{R}^{n_p} \rightarrow \mathbb{R}^{n_p}$ is a translation vector, where $n_p$ is the position dimension. The set $\mathbb{V}_{0}$ is described as $\mathbb{V}_{0}:=\left\{y\in \mathbb{R}^{n}:A_{0}y\leq b_{0}\right\}$, where $A_{0}$ and $b_{0}$ are depends on the length $L$ and width $W$ of a car.

In a given parking environment, we assume that there are $N_{\mathbb{O}}$ static obstacles denoted by:
\begin{equation}
    \mathbb{O}_{i} := \left\{ y\in \mathcal{R}^{n}: A^{\mathbb{O}}_iy \leq b^{\mathbb{O}}_i\right\},
\end{equation}
where $A^{\mathbb{O}}_i \in \mathbb{R}^{s\times n}$, $b^{\mathbb{O}}_i\in \mathbb{R}^{s}$, where $s$ represents the faces of an obstacle. 
$\mathbb{O}_{i}, i\in \left\{1,...,N_{\mathbb{O}}\right\}$ are assumed to be convex compact sets and non-empty.
Usually, we only need to consider the obstacles within the grid $\Omega_\mathbf{z}$, denoted by $\mathbb{O}^{\Omega}_i$.

After we construct the polytope representation of $\mathbb{V}_k$ and $\mathbb{O}^{\Omega}_i$, we can formulate a QP problem to get the minimal signed distance $d^{\mathbf{z}_k}_i$ between each pair $(\mathbb{V}_k, \mathbb{O}^{\Omega}_i)$. 
More details of the computation are referred to~\cite{thirugnanam2022duality}.
The safe set contains $\mathbf{z}_k$ with a positive signed distance denote by
\begin{equation}
    \mathcal{C}_{i} := \left\{\mathbf{z}_{k}\in \Omega_{\mathbf{z}}: d_{i}^{\mathbf{z}_{k}} > 0\right\},
\end{equation}
where $\mathcal{C}_i$ is the safe set considering $\mathbb{O}^{\Omega}_i$, the complete safe set is $\mathcal{C} := \bigcap ^{N_{\mathbb{O}}}_{i=1} \mathcal{C}_{i}$.

\subsection{Safe Reachable Set}
The BRT $\mathcal{G}$ guarantees that the states in the tube will reach the parking spot while taking into account the system dynamics. 
On the other hand, the safe set $\mathcal{C}$ guarantees that the states are collision-free. 
The values of states $\mathbf{z}_k$  are negative within $\mathcal{G}$ while and zeros on the boundary $\partial {\mathcal{G}}$.
To obtain the safe reachable set, denoted by $\mathcal{S}$, we perform an intersection operation of the Backward reachable tube and the safe set, i.e., $\mathcal{S} := \mathcal{G}\cap \mathcal{C}$. This ensures that the resulting set contains only states that are both kinematically feasible and collision-free, which is critical for safe and efficient parking.

\revised{\subsection{Connected States Sampling}}
\revised{In a given parking scenario, we compute the safe reachable set $\mathcal{S}$. 
The connected state is acquired by uniform sampling as below
\begin{equation}
    \mathbf{z}_{k}^\mathcal{S}=\left\{(x_k,y_k,\theta_k)|x_k\geq x_g, y_k\geq y_g, (x_k, y_k, \theta_k) \in \mathcal{S}\right\},
\end{equation}
Where $x_g, y_g$ are the coordinates of the parking goal pose.
The number of connected states is 20.

The target parking pose often dictates the nature of the parking scenario by determining which states are connected. We've formulated a safe reachable set that represents the intersection of BRT and the safe set. This approach helps to eliminate undesirable states that may occur in the BRT when time is minimal, which can result in the connected states being too proximate to the parking goal. An intuitive way to understand this concept is by visualizing the undesirability of having a person standing in the parking spot directing us – a situation that could be complicated by the presence of other obstacles. Rather, it is more feasible to have guidance from people positioned outside and around the parking spot.
The quantity of sampling states is user-dependent.

While the safe set $\mathcal{C}$, is devoid of collision-prone states, it does not take into account their viability due to potential obstruction from obstacles between the parking spot and the states within the safe set. In any particular parking scenario, it is not optimal to sample connected states that, although safe, could pose difficulty or inefficiency in directing the search due to their lack of kinematic viability. This is an important aspect captured by the safe reachable set $\mathcal{S}$, and not considered in the safe set.

}
\section{online bidirectional search}\label{sec: online}
In this section, we explore an A*-based variant approach to achieve fast searching with the help of connected points as shown in Fig.~\ref{fig:general_framework}.

\subsection{Search Strategy}
To improve the search speed in a given parking environment, we employ both forward search and backward search to connect states. Bidirectional variants of planning algorithms are frequently employed in problems with challenging regions, such as narrow environments or high-dimensional configuration spaces~\cite{lavalle2006planningalgorithm}. In the case of a parking planning problem, where the parking spot is often surrounded by parked vehicles and obstacles, a bidirectional search strategy can be highly effective. The search logic for our approach is illustrated in Algorithm~\ref{alg:BA*_search_logic}.

A parking path, denoted by $P(\mathbf{z}_0, \mathbf{z}_g)$, consists of two parts: the path searched by node expansion, represented by $\mathcal{A}(\cdot, \cdot)$, and the path found by Reeds-Shepp curves, represented by $\mathcal{RS}(\cdot, \cdot)$. The size of a set is denoted by $\left|\cdot\right|$. The commonly used formulation for a parking planning problem is given by:
\begin{equation}\label{eq:HA*}
P(\mathbf{z}_{0}, \mathbf{z}_{g}) = \mathcal{A}(\mathbf{z}_{0}, \mathbf{z}_{\text{best}}) \oplus \mathcal{RS}(\mathbf{z}_{\text{best}}, \mathbf{z}_{g})
,\end{equation}
where $\oplus$ denotes the \emph{CombinePath} operation. 

We can use \eqref{eq:HA*} to describe a common solution to a parking planning problem, the HA* algorithm can be computationally expensive due to its reliance on inefficient node expansion, which can lead to a large set size of $\left|\mathcal{A}\right|$.
Heuristic bidirectional search constructs a path
\begin{equation}
P^{\textup{BA*}}_{(\mathbf{z}_{0}, \mathbf{z}_{g})} = P(\mathbf{z}_0 \rightarrow \mathbf{z}^\mathcal{S}_i) \oplus P(\mathbf{z}^\mathcal{S}_i \leftarrow \mathbf{z}_g).
\end{equation}
In this strategy, the possibility of a bigger $\left |\mathcal{RS}\right |$ increases because the \emph{RSExpansion} to connected state $\mathbf{z}^{\mathcal{S}}_{i}$ is relatively easier.
Single directional search strategy used in HA*~\cite{dolgov2008hybridA} and~\cite{aine2016multi} can not decrease $\left|\mathcal{A}\right|$ efficiently.

\begin{remark}
The goal of HJBA* is to reduce the size of $\left|\mathcal{A}\right|$ and increase the size of $\left|\mathcal{RS}\right|$ with the help of connected states $\mathbf{z^{\mathcal{S}}_{i}}$. 
For instance, in the case where $\mathbb{O} = \emptyset$, we always have $\left|\mathcal{A}\right| = 0 $, and the path $P$ is equivalent to $\mathcal{RS}$ curves. 
In this case, the planning time equals to analytical computation time which is very fast and negligible.
This feature is fully utilized within our framework. 
\end{remark}

\begin{algorithm}[!t]
    \SetKwInOut{KwOut}{Return}
	\caption{BA*:Search Strategy}
	\label{alg:BA*_search_logic}
	\KwIn{$\mathbb{O}, \mathbf{z}_{0}, \mathbf{z}_{g}, \mathbf{z}^\mathcal{S}_i$}
	$Q^{\textup{F}}_{\textup{open}} \gets \emptyset, Q^{\textup{F}}_{\textup{close}} \gets \emptyset, flag^{\textup{F}} \gets $ \textbf{\upshape False}\;
	$Q^{\textup{B}}_{\textup{open}} \gets \emptyset, Q^{\textup{B}}_{\textup{close}} \gets \emptyset, flag^{\textup{B}} \gets $ \textbf{\upshape False}\;
	$ Q^{\textup{F}}_{\textup{open}} \gets \text{\upshape Initialize}(\mathbf{z}_{0})$, $ Q^{\textup{B}}_{\textup{open}} \gets \text{\upshape Initialize}(\mathbf{z}_{g})$\; 
	\While{$Q_{\textup{open}}^{\textup{F}}\neq \emptyset$ \textbf{\upshape and} $Q_{\textup{open}}^{\textup{B}}\neq \emptyset$}{
		\If{$flag^{\textup{F}}$ \textbf{\upshape and} $flag^{\textup{B}}$}
		{\textbf{\upshape break}\;}
		\If{\textbf{\upshape not} $flag^{\textup{F}}$}{$\mathbf{z}^{\textup{F}}_{\text{best}}=Q^{\textup{F}}_{\textup{open}}$.pop: $c(\mathbf{z}^{\textup{F}}_{\text{best}}) < c(\mathbf{z}^{\textup{F}}), \forall \mathbf{z}^{\textup{F}}\in \mathcal{F}$\par 
		$(flag^{\textup{F}}, \mathcal{RS}(\mathbf{z}^{\textup{F}}, \mathbf{z}^\mathcal{S}_i)) \gets \text{\upshape RSExpansion}(\mathbf{z}^{\textup{F}}, \mathbf{z}^\mathcal{S}_i)$\par
		$\text{\upshape ForwardExpand}(\mathbf{z}^{\textup{F}}_{\text{best}}, \mathcal{F}$, $Q^{\textup{F}}_{\textup{open}}, Q^{\textup{F}}_{\textup{close}})$}
		\If{\textbf{\upshape not} $flag^{\textup{B}}$}{	$\mathbf{z}^{\textup{B}}_{\text{best}}=Q^{\textup{B}}_{\textup{open}}$.pop: $c(\mathbf{z}^{\textup{B}}_{\text{best}}) < c(\mathbf{z}^{\textup{B}}), \forall \mathbf{z}^{\textup{B}}\in \mathcal{B}$\par
		$(flag^{\textup{B}}, \mathcal{RS}(\mathbf{z}^{\textup{B}}, \mathbf{z}^\mathcal{S}_i)) \gets \text{\upshape RSExpansion}(\mathbf{z}^{\textup{B}}, \mathbf{z}^\mathcal{S}_i)$\par
		$\text{\upshape BackwardExpand}(\mathbf{z}^{\textup{B}}_{\text{best}}, \mathcal{B}$, $Q^{\textup{B}}_{\textup{open}}, Q^{\textup{B}}_{\textup{close}})$}
	}
	$P(\mathbf{z}_0 \rightarrow \mathbf{z}^\mathcal{S}_i)$ = GetPath$(Q^{\textup{F}}_{\textup{close}},\mathcal{RS}(\mathbf{z}^{\textup{F}}, \mathbf{z}^\mathcal{S}_i, \mathbf{z}_{0}))$\;
	$P(\mathbf{z}^\mathcal{S}_i \leftarrow \mathbf{z}_g)$ = GetPath$(Q^{\textup{B}}_{\textup{close}},\mathcal{RS}(\mathbf{z}^{\textup{B}}, \mathbf{z}^\mathcal{S}_i, \mathbf{z}_{g}))$\;
	$P(\mathbf{z}_{0}, \mathbf{z}_{g})$ = CombinePath$(P(\mathbf{z}_0 \rightarrow \mathbf{z}^\mathcal{S}_i),P(\mathbf{z}^\mathcal{S}_i \leftarrow \mathbf{z}_g) )$\;
	
 	\KwOut{$P(\mathbf{z}_{0}, \mathbf{z}_{g})$}
\end{algorithm}

\subsection{Node Expansion}
We use different node expansions in \emph{ForwardExpand} and \emph{BackwardExpand}.
The motion primitives for bidirectional search are defined as:
\begin{equation}
\begin{split}
    \mathcal{F}&= \left\{ f(\mathbf{z}_{\text{best}}, u_k)d_{k}: u_{k}\in \mathbf{U}_{\textup{F}} \right\},\\
    \mathcal{B}&= \left\{ f(\mathbf{z}_{\text{best}}, u_k)d_{k}: u_{k}\in \mathbf{U}_{\textup{B}} \right\},
\end{split}
\end{equation}
where $f(\cdot)$ is the system dynamics~\ref{eq:system_dynamics}.
For convenience, let $\mathcal{F}=\mathcal{F}^{\textup{F}} \cup \mathcal{F}^{\textup{B}}$, where $\mathcal{F}^{\textup{F}}$ contains the forward node expansion by the action $u_k\in \mathbf{U}_{\textup{F}}$ and $d_{k}=1$, analogously to $\mathcal{F}^{\textup{B}}$ with $d_{k}=-1$.

\begin{figure}
\centering
\includegraphics[width=1\linewidth]{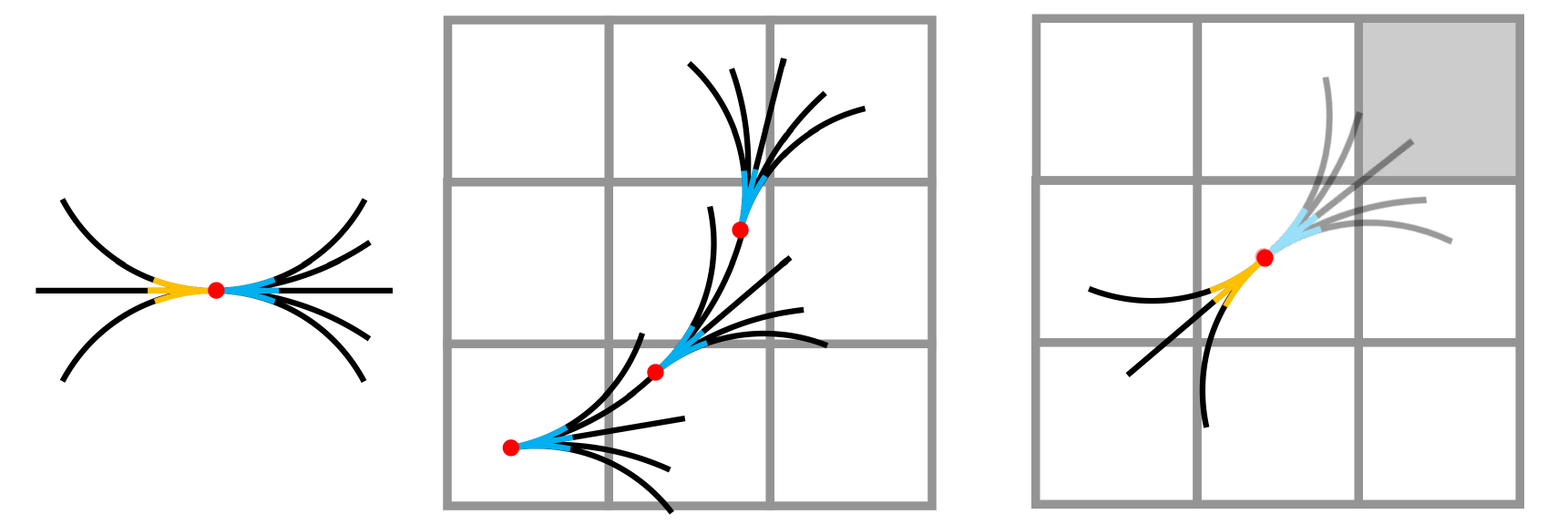}
\caption{Motion Primitives and expansion mode of the search algorithm.}
\label{fig:motion_premitives}
\end{figure}

The node expansion of \emph{forward search} is always towards the connected state $\mathbf{z}^\mathcal{S}_i$.
If all forward expansions $\mathcal{F}^{\textup{F}}$ failed, the maximum turning angle of backward expansions $\mathcal{F}^{\textup{B}}$ are used to keep searching as illustrated in Fig.~\ref{fig:motion_premitives}.
The node expansion for \emph{ForwardExpand} is detailed in algorithm~\ref{alg:BA*_NodeExpand}.
The node expansion of \emph{BackwardExpand} stretches out in both directions because the parking spot is often narrow and a smaller motion resolution $r^{\textup{B}} < r^{\textup{F}}$ is preferred. 

\begin{remark}
    The proposed node expansion strategy may lead to more aggressive expansion for traditional A*-based search algorithms due to its focus on backward expansion only when close to obstacles, which can result in increased search time. However, the use of connected states as heuristic points reduces the reliance on backward node expansion to find  $\mathbf{z}_{\textup{best}}$.
\end{remark}

\begin{algorithm}[ht]
    \SetKwInOut{KwOut}{Return}
	\caption{BA*:NodeExpand}
	\label{alg:BA*_NodeExpand}
	\KwIn{$\mathcal{F}, \mathbf{z}^{\textup{F}}_{\text{best}}, Q^{\textup{F}}_{\textup{open}}, Q^{\textup{F}}_{\textup{close}}$}
	{\While{$k \leq \left | \mathcal{M}\right |$}
	{$\mathbf{z}_{k}\gets$ GetNode($\mathcal{F}^{\textup{F}}$, $\mathbf{z}_{\text{best}}$)\par
	CollisionCheck($\mathbf{z}_{k}$)\par
	\If{$k= \left |\mathcal{F}^{\textup{F}} \right |$}{break}
	\If{$\mathbf{z}_{k}\in Q^{\textup{F}}_{\textup{close}}$}{continue}
	\uIf{$\mathbf{z}_{k}\notin Q_{\textup{open}}$}{$Q_{\textup{open}}\gets Q_{\textup{open}}\cup \mathbf{z}_{k}$}
	\ElseIf{$c(Q^{\textup{F}}_{\textup{open}}(\mathbf{z}_{k})) > c(\mathbf{z}_{k})$}{$Q^{\textup{F}}_{\textup{open}}(\mathbf{z}_{k})\gets \mathbf{z}_{k}$}
	}}
 	\KwOut{$Q^{\textup{F}}_{\textup{open}}$, $Q^{\textup{F}}_{\textup{close}}$}
\end{algorithm}

\subsection{Cost Function}
The node cost in HJBA* is defined as
\[
c(\mathbf z)=g(\mathbf z_0,\mathbf z)+h(\mathbf z,\mathbf z_i^S),
\]
where $g(\mathbf z_0,\mathbf z)$ is the cost-to-come from the initial state to the current node, and $h(\mathbf z,\mathbf z_i^S)$ is the heuristic cost-to-go from the current node to the connected state $\mathbf z_i^S$ used by the current search branch.

In our implementation, the cost-to-come $g(\mathbf z_0,\mathbf z)$ is computed as the accumulated length of the motion primitives. Different from conventional Hybrid A* methods, we do not introduce additional penalty terms for steering changes, backward motions, or direction switches. Although such penalties can bias the search toward certain maneuver preferences, they usually require scenario-dependent tuning and may lead to unstable behavior across different parking layouts.

The reason that a simple heuristic is sufficient in HJBA* is that the dominant search guidance is not provided by the heuristic alone, but by the connected states sampled from the safe reachable set together with the node-expansion strategy that explicitly steers the search toward these connected states. Therefore, the heuristic only needs to provide a lightweight and stable local estimate for node ordering in the open list.
Accordingly, we use the diagonal distance in the discretized $(x,y)$ plane to the connected state $\mathbf z_i^S$ as the heuristic:
\[
h(\mathbf z,\mathbf z_i^S)
=
D\big((x,y),(x_i^S,y_i^S)\big),
\]
where $D(\cdot,\cdot)$ denotes the diagonal distance. This heuristic is parameter-free and computationally inexpensive.

Moreover, the diagonal heuristic is consistent. Since $g$ is computed from the actual motion primitive length, the one-step transition cost between two neighboring nodes is no smaller than their translational displacement in the $(x,y)$ plane. Meanwhile, $h(\mathbf z,\mathbf z_i^S)$ only measures the remaining planar distance and ignores additional costs caused by heading mismatch and obstacle avoidance; therefore, it does not overestimate the true remaining cost. Because the diagonal distance satisfies the triangle inequality, for any adjacent nodes $\mathbf z$ and $\mathbf z'$ we have
\[
h(\mathbf z,\mathbf z_i^S)\le c(\mathbf z,\mathbf z')+h(\mathbf z',\mathbf z_i^S),
\]
which shows that the heuristic is consistent.

\subsection{Continuity of Planned Parking Paths}
In this section, we provide evidence to support the claim that the planned path is continuous.

\begin{figure}[!h]
  \centering
    \begin{subfigure}{\linewidth}
        \includegraphics[width=\linewidth]{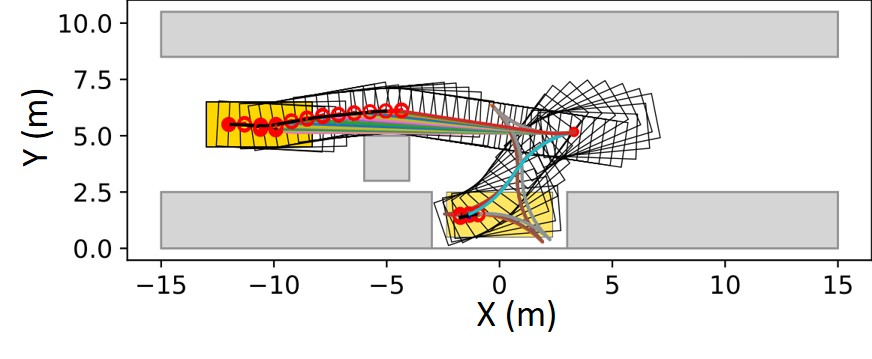}
        \caption{HJBA*: Expansions.}
    \label{fig:vertial_conti}
  \end{subfigure}

    \begin{subfigure}{\linewidth}
        \includegraphics[width=\linewidth]{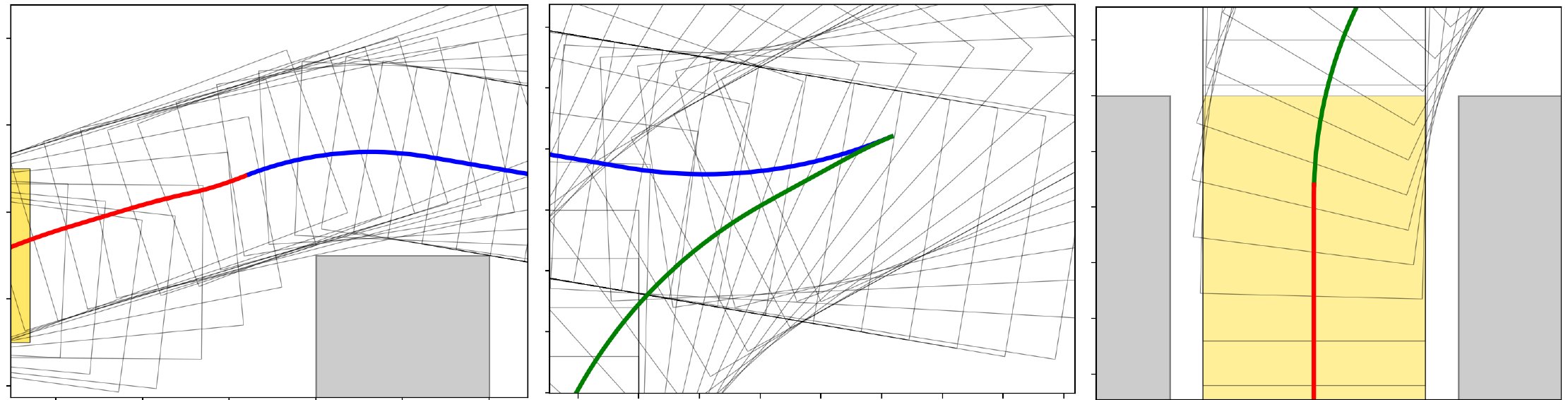}
        \caption{The details of connection regions.}
    \label{fig:vertical_big}
  \end{subfigure}
  
  \begin{subfigure}{\linewidth}
    \centering
        \includegraphics[width=\linewidth]{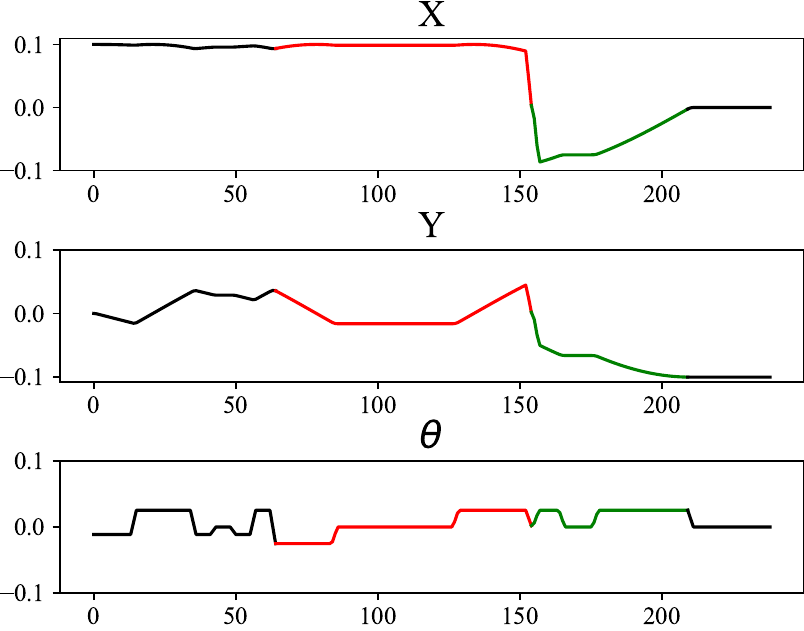}
        \caption{The continuity of planned parking path.}
    \label{fig:vertical_grad}
  \end{subfigure} 
  \caption{The highly efficient searching strategy.}
\end{figure}

Figure~\ref{fig:vertial_conti} visually demonstrates the continuity of the path in an intuitive manner. The connected segments of the path are formed by the $\mathcal{RS}$ curves, which ensure a smooth transition between different parts of the path. To provide a clearer perspective of the connected region, Fig.~\ref{fig:vertical_big} showcases the extent of the continuous path.

It is important to note that the parking path exhibits a single cusp at the point where the direction changes. This cusp represents the location of the direction change in the path. To further analyze the continuity, Fig.~\ref{fig:vertical_grad} displays the computed derivatives of the path along the $x$, $y$ and curvature information related to $\theta$. The results confirm the continuity of the path.

Overall, the visual demonstration and numerical derivative analysis provide substantial evidence supporting the continuity of the planned parking paths in our approach.

\subsection{Demonstration of Motion Primitives and Search Strategy}
One notable advantage of our HJBA* algorithm is its versatility in handling various parking scenarios without the need for parameter adjustments. In contrast, traditional HA*-based algorithms often require parameter tuning to achieve optimal results.

\begin{figure}[!h]
  \centering
    \begin{subfigure}{\linewidth}
        \includegraphics[width=\linewidth]{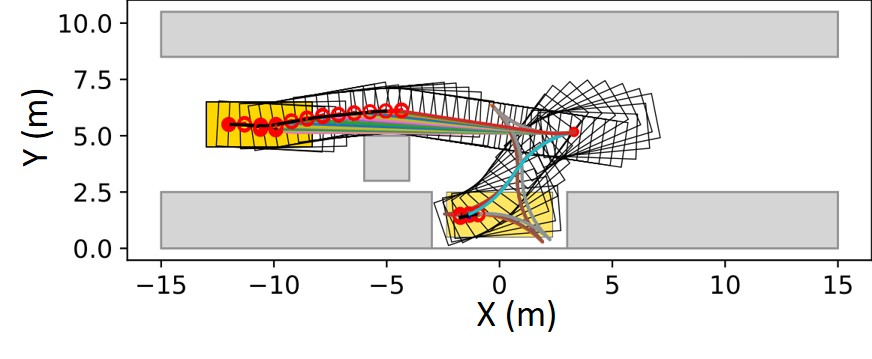}
        \caption{The parallel parking showcase}
    \label{fig:param_a}
  \end{subfigure}
  
  \begin{subfigure}{\linewidth}
    \centering
        \includegraphics[width=\linewidth]{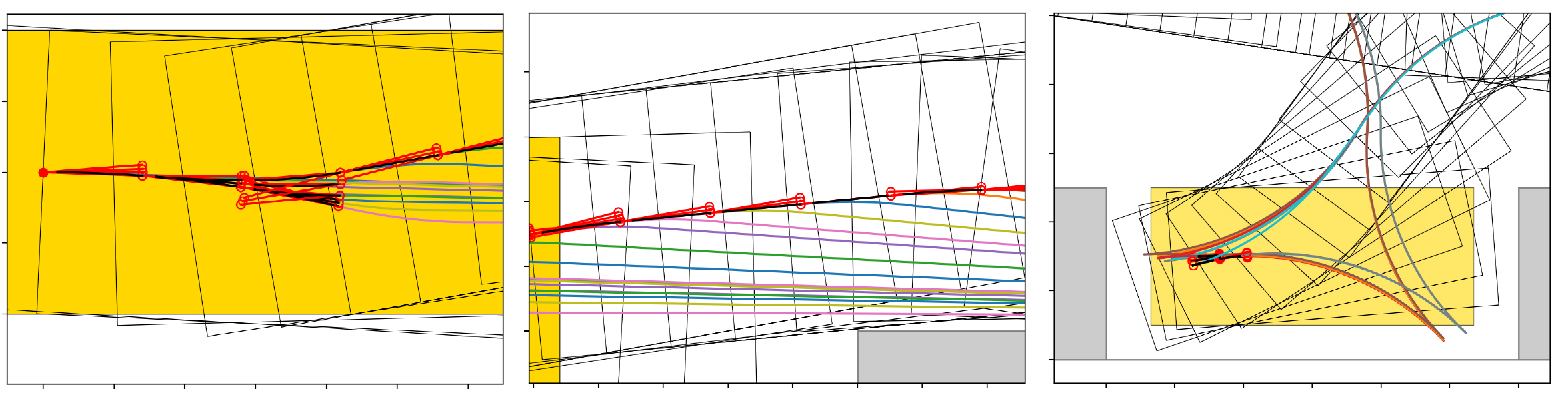}
        \caption{Motion primitives and search strategies}
    \label{fig:param_b}
  \end{subfigure} 
  \caption{The demonstration of the searching strategy without parameters tuning.}
\end{figure}

To showcase the benefits of our designed search and node expansion strategy, as well as the utilization of connected states $\mathbf{z}^{\mathcal{S}}_k$ without parameter tuning, we present a parallel parking example in Fig.~\ref{fig:param_a}. Our search tree consistently expands nodes towards the connected states, eliminating the need for backward node expansion. Consequently, we eliminate the need for penalty parameters to discourage backward movements. This robustness is maintained even if the start pose is on the opposite side, as the vehicle will still expand towards the connected states. Hence, there is no requirement for fine-tuning penalty parameters for switching directions.

In situations where the search becomes obstructed by an obstacle, as depicted in Fig.~\ref{fig:param_b}, our algorithm adapts by expanding nodes in the backward direction. Once a collision-free path is identified, the search continues towards the connected states. The backward search stage often involves obstacles and narrow parking spaces, making different motion resolutions and search strategies crucial. In these scenarios, our algorithm employs a double-direction motion primitives in the backward search phase, as it significantly aids in discovering collision-free paths in narrow environments.

The demonstrated examples effectively highlight the effectiveness of our motion primitives and search strategy. By leveraging connected states and employing appropriate search strategies, our approach eliminates the need for parameter adjustments while providing robust and efficient planning in various parking scenarios.

\subsection{Distinguishing Features from Other Bidirectional A* Approaches}
In search-based algorithms, such as A*-based methods~\cite{adabala2020multi}, and sampling-based algorithms, such as RRT-based techniques~\cite{jhang2020forward}, bidirectional search strategies have been widely employed. However, our algorithm differs from previous bidirectional A* approaches by introducing the concept of connected states.

Unlike traditional bidirectional search strategies where the forward search targets the goal pose and the backward search targets the initial pose, our algorithm guides both searches towards connected states instead of merging in a specific region. This distinction is particularly beneficial in scenarios where the target is surrounded by obstacles, such as clustered and narrow parking scenarios or complex environments. In such cases, conventional bidirectional search algorithms tend to exhibit inefficiency.

Moreover, our approach takes into account the efficiency of $\mathcal{RS}$ curves, leveraging the safe reachable set to provide guidance to the connected states. This eliminates the need to fine-tune the hyperparameters of heuristic-based search algorithms. Furthermore, our motion primitives facilitate efficient expansion of nodes towards the connected states, contributing to the overall effectiveness of the approach.

\begin{remark}
While our approach employs bidirectional search and $\mathcal{RS}$ curves to generate a complete parking path, it is important to note that the path remains continuous throughout. 
We will provide a detailed demonstration in a later section of the paper.
\end{remark}

\section{Simulation and experiment results}
\label{sec: results}
\subsection{Simulation setup}
All simulations are conducted on a laptop running Ubuntu 22.04, equipped with an Intel Core i7-13650HX CPU. The ideal parking path planning algorithm should be \emph{versatile} enough to handle various parking scenarios, including both common and exceptional cases, \emph{robust} in accommodating different parking tasks (e.g., from arbitrary initial and goal poses), \emph{stable} to minimize solution variance, and \emph{computationally efficient} to provide fast solutions.
To demonstrate the exceptional performance of our proposed HJBA* algorithm based on these criteria, we first validate it with a set of simple examples. Subsequently, we evaluate our parking path planner through extensive simulations, comparing it with several state-of-the-art planning algorithms: HA*~\cite{dolgov2008hybridA}, Multi-heuristic Hybrid A (MHHA*)\cite{huang2022search}, Fault-Tolerant Hybrid A* (FTHA*)\cite{bai2022irregularly}, and Mirroring the Parking Target (MPT)\cite{Jia2024mirror}. 
The parking spot size and parameters used in the simulations are listed in Tab.~\ref{tab:parameters}, except for those involving irregular parking scenarios.

\begin{table}[]
    \centering
    \caption{\revised{Setup of the simulation parameter associated with algorithms and vehicles}}
    \label{tab:parameters}
    \revised{
\begin{tabular}{l|l|l}
\hline

Setup                                                                   & Parameter                    & Values                      \\ \hline
\multirow{7}{*}{\begin{tabular}[c]{@{}l@{}}HA*\\ \\ MHHA*\end{tabular}} & Grid size                     & 0.5 m                       \\
                                                                        & Yaw resolution                & 5$^{\circ}$                 \\
                                                                        & Motion resolution             & 0.1 m                       \\
                                                                        & Switch back penalty           & 5.0                         \\
                                                                        & Back penalty                  & 5.0                         \\
                                                                        & Steer change penalty          & 5.0                         \\
                                                                        & Steer penalty                 & 5.0                         \\ \hline
\multirow{4}{*}{HJBA*}                                                  & Grid size for forward search  & 0.5 m                       \\
                                                                        & Grid size for backward search & 0.3 m                       \\
                                                                        & Yaw resolution                & 5$^{\circ}$                 \\
                                                                        & Parallel computation threads  & 12                          \\ \hline
\multirow{4}{*}{Vehicle}                                                & Width of the vehicle          & 2.0 m                       \\
                                                                        & Length of the vehicle         & 4.7 m                       \\
                                                                        & Max steering angle            & 0.6 rad                     \\
                                                                        & Wheelbase                     & 2.7 m                       \\ \hline
\multirow{3}{*}{Parking spot size}                                      & Perpendicular$(w,d,h,\alpha)$ & 2.6, 5.0, 6.0, $90^{\circ}$ \\
                                                                        & Angle $(w,d,h,\alpha)$        & 2.6, 5.0, 6.0, $45^{\circ}$ \\
                                                                        & Parallel$(w,d,h,\alpha)$      & 6.0, 2.5, 6.0, $90^{\circ}$ \\ \hline
\end{tabular}}
\end{table}
\subsection{Validation and Demonstration}
In this section, we validate and demonstrate the workflow of HJBA* through three common parking scenarios.

\subsubsection{Safe reachable set demonstration}
\begin{figure}[t]
  \centering
  \begin{subfigure}{\linewidth}
    \centering
    \includegraphics[width=\linewidth]{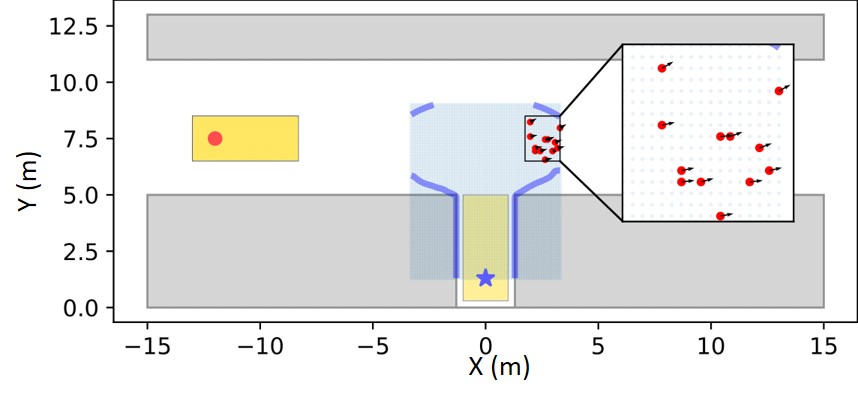}
    \caption{High-level of HJBA*: Perpendicular parking}
    \label{subfig:perpendicular_example_layer1}
  \end{subfigure}

  \begin{subfigure}{\linewidth}
    \centering
    \includegraphics[width=\linewidth]{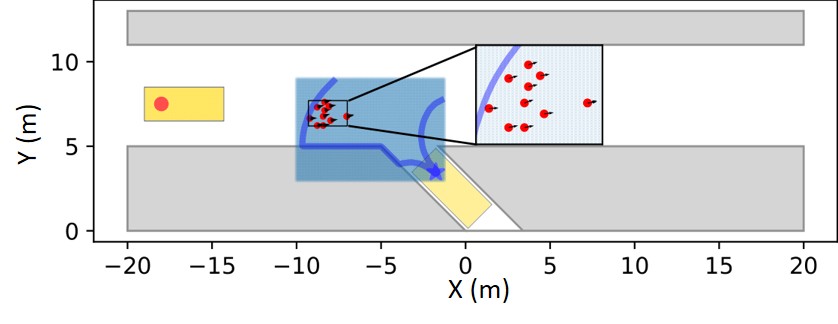}
    \caption{High-level of HJBA*: Angle Parking}
    \label{subfig:angle_example_layer1}
  \end{subfigure}  
  
    \begin{subfigure}{\linewidth}
    \centering
    \includegraphics[width=\linewidth]{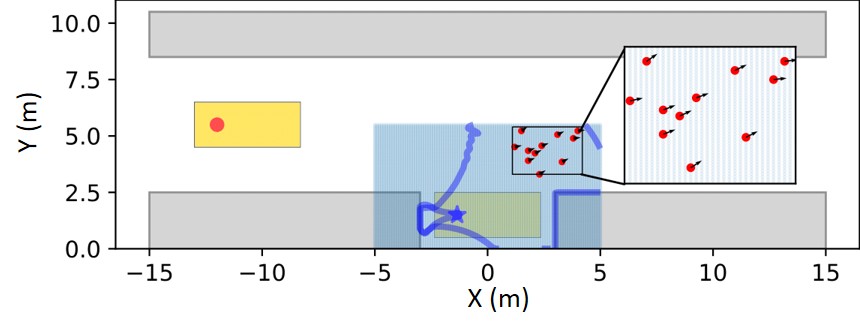}
    \caption{High-level of HJBA*: Parallel Parking}
    \label{subfig:parallel_example_layer1}
  \end{subfigure}  
  \caption{Projection of the sampling tube onto a 2D space, with randomly sampled states as examples. 
  The obstacles are patched gray.
  The red circle is the initial pose and blue star is the parking goal pose.
  The light blue region is the grid for BRT computation. 
  The solid blue lines are the bound of $\mathcal{G}$ projected in the 2-dimensional plane and there are twelve random sampling states with directions shown as red dots with black arrows.}
\end{figure}

The first scenario we consider is a perpendicular parking situation in a confined space, illustrated in Fig.~\ref{subfig:perpendicular_example_layer1}. The gri $\Omega_{\mathbf{z}}$ has dimensions of $61 \times 61 \times 61$, and the starting pose is located at $[-12, 7.5, 0.0]$, while the goal pose is at $[0, 1.3, \pi/2]$. In this context, the safe reachable set refers to the area that drivers must first traverse when parking, and the connected states represent various choices available to drivers during the parking process.

The second example we examine involves angle parking, shown in Fig.~\ref{subfig:angle_example_layer1}. The $\Omega_{\mathbf{z}}$ used in this scenario has dimensions of $101\times 101 \times 101$, and the starting pose is situated at $[-18, 7.5, 0.0]$, while the goal pose is located at $[-1.7, 3.5, -\pi/4 ]$. 
In this context, drivers must initiate their parking maneuvers before approaching the parking spot, and as a result, the sampling states are primarily concentrated in the left area. Additionally, it is possible for the initial parking pose to be located in the right area of the parking spot since the sampling states can guide drivers to move back towards the parking spot.

The third scenario we consider is a parallel parking situation, as shown in Fig.~\ref{subfig:parallel_example_layer1}. The computation grid used in this scenario has dimensions of $101\times 101 \times 101$, and the starting pose is located at $[-12, 5.0, 0.0]$, while the goal pose is situated at $[-1.35, 1.5, 0.0 ]$. Parallel parking is generally more complex than the other types of parking since, in narrow parallel parking spots, the ego car needs to move back and forth several times to leave the parking spot. In contrast, adjustments to the position in perpendicular or angle parking mostly depend on the size $h$ of the driving space. In narrow parallel parking spaces, it is often necessary for the car to move back first to create more space for forward movement in the subsequent stage. To address this issue, we compute $\mathcal{S}$ for the first stage, which is used as the goal parking set for the next stage. The connected states are randomly distributed in the right area, as the car cannot park from the rear side of the parking spot due to obstacles.

\subsubsection{The cooperation of bidirectional search and safe reachable set}
\begin{figure}[!ht]
  \centering
  \begin{subfigure}{\linewidth}
    \centering
    \includegraphics[width=\linewidth]{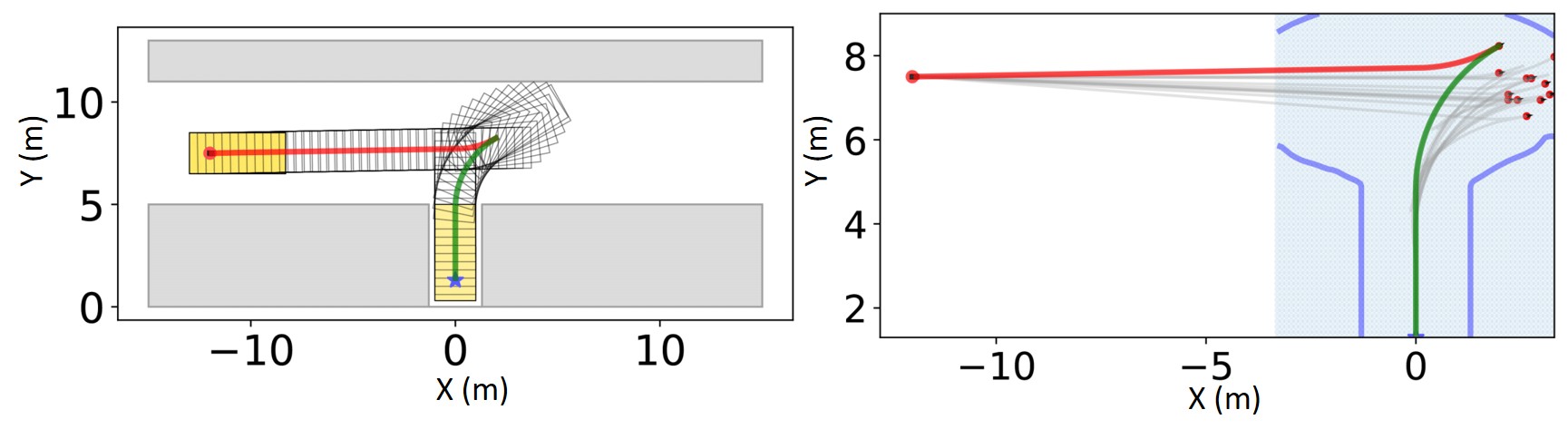}
    \caption{Perpendicular Parking}
    \label{subfig:perpendicular_running_example}
  \end{subfigure}

  \begin{subfigure}{\linewidth}
    \centering
    \includegraphics[width=\linewidth]{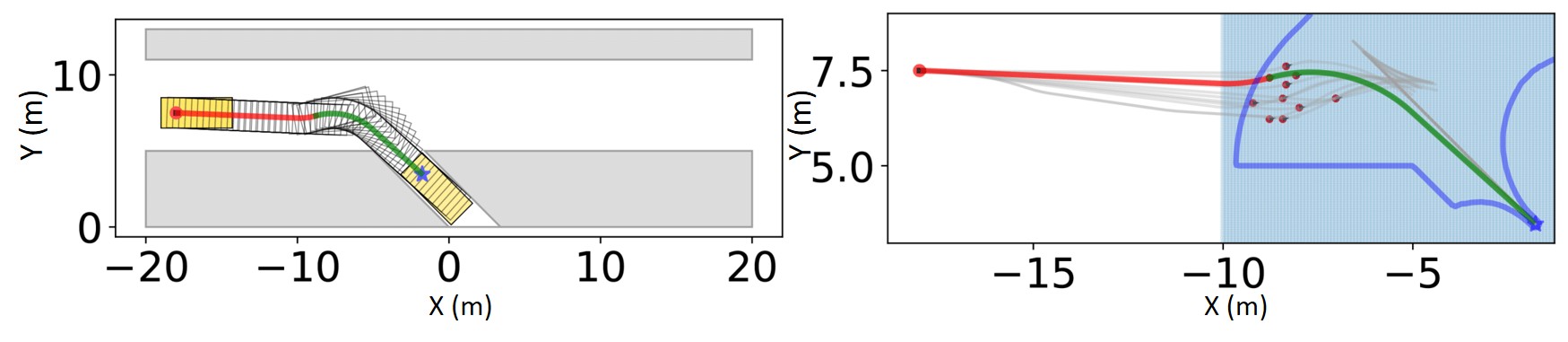}
    \caption{Angle Parking}
  \end{subfigure}  
  
    \begin{subfigure}{\linewidth}
    \centering
    \includegraphics[width=\linewidth]{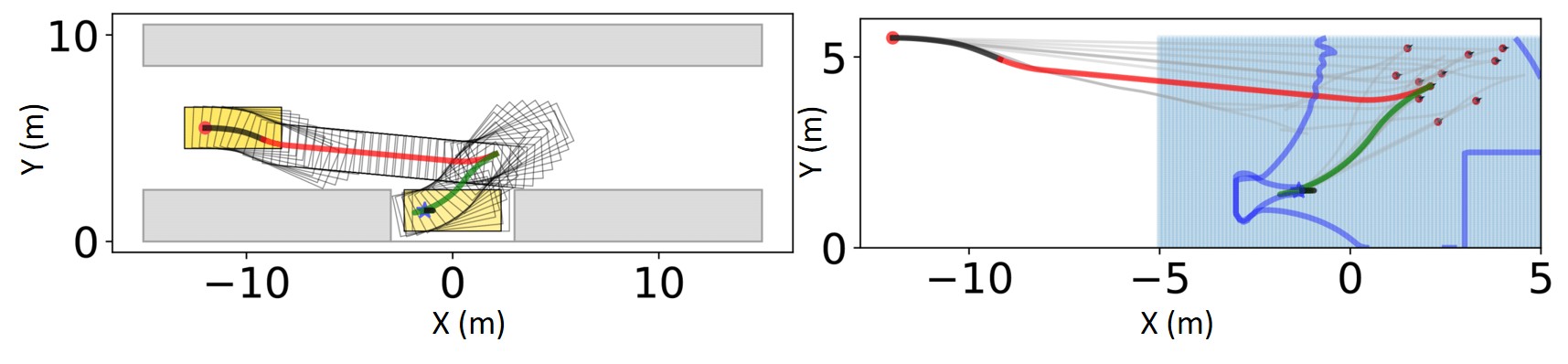}
    \caption{Parallel Parking}
  \end{subfigure}  
  \caption{The illustration of HJBA* in different parking scenarios. 
  Left: The parking path $P(\mathbf{z}_{0}, \mathbf{z}_{g})$ is composed of line segments corresponding to the right side.
  Right:The red segment denotes $\mathcal{RS}(\mathbf{z}^{\textup{F}}_{\text{best}}, \mathbf{z}^\mathcal{S}_i)$, the black segment denotes $\mathcal{A}(\mathbf{z}_{0}, \mathbf{z}^{\textup{F}}_{\text{best}})$ and $\mathcal{A}(\mathbf{z}_{g}, \mathbf{z}^{\textup{B}}_{\text{best}})$, and the green segment denotes $\mathcal{RS}(\mathbf{z}^{\textup{B}}_{\text{best}}, \mathbf{z}^\mathcal{S}_i)$.} 
  \label{fig:running_example_for_low_level}
\end{figure}

We showcase our search-based algorithm in three typical parking scenarios in Fig.~\ref{fig:running_example_for_low_level}. 
As evident from the sub-figures on the right-hand side, each connected state $\mathbf{z}_k\in\mathcal{S}$ is connected to the starting and ending points by a gray parking path $P(\mathbf{z}_{0}, \mathbf{z}_g)$, which is obtained through parallel computation. 
Users can choose the preferred path with the shortest length or computation time. Regardless of the parking scenario, the parking path primarily consists of RS curves and an A*-based expansion segment. 
Similar to our previous design objective, the length of the RS curves is increased, while the length of the graph-based expansion is reduced, thereby significantly improving the computation speed.

\subsubsection{The HJBA* performance demonstration}
\begin{figure}[!h]
  \centering
    \begin{subfigure}{\linewidth}
        \includegraphics[width=\linewidth]{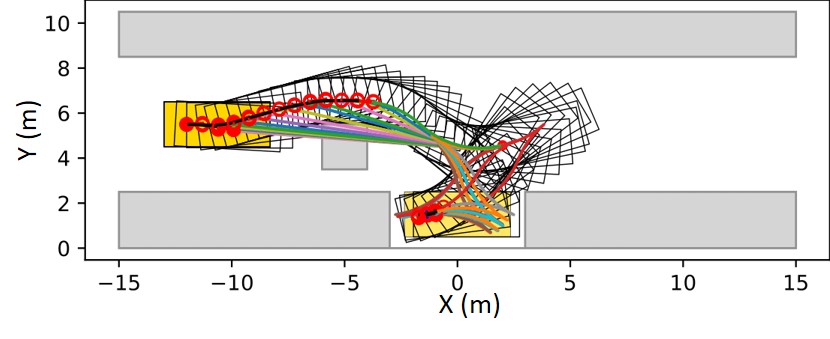}
        \caption{ HJBA*: Expansions}
    \label{fig:bi_rs}
  \end{subfigure}
    \begin{subfigure}{\linewidth}
    \centering
        \includegraphics[width=\linewidth]{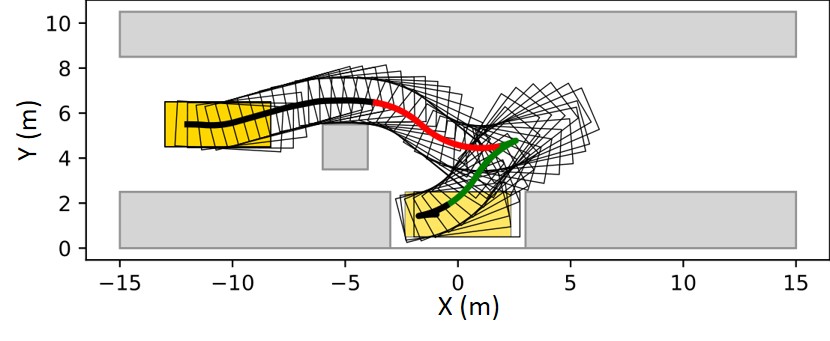}
        \caption{HJBA*: Results}
    \label{fig:bi_rs_show}
  \end{subfigure} 
  
    \begin{subfigure}{\linewidth}
    \centering
        \includegraphics[width=\linewidth]{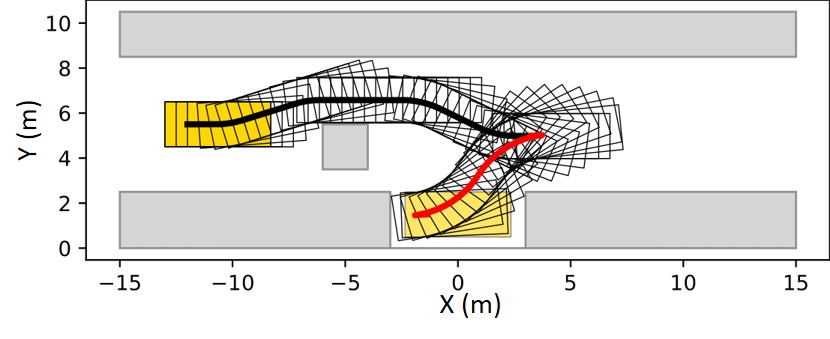}
        \caption{HA*: Results}
    \label{fig:ori_rs_show}
  \end{subfigure}

  \caption{The demonstration for search and comparison of HJBA* and HA* in one case.
  On left-hand side (a) and (c),
  the colorful curves are $\mathcal{RS}$ curves and the red circles are visualization of expansion nodes during the searching process.
  On right-hand side (b) and (d), the black segment is collision-free $\mathcal{A}$, the red and green segment are collision-free $\mathcal{RS}$.}
  \label{fig:comparison}
  \end{figure}

To showcase the superiority of our HJBA*, we present an example illustrated in Fig.~\ref{fig:comparison}.
Our HJBA* and HA* algorithms start from the same initial state $\mathbf{z}_0 = [-12, 5.5, 0, 0]$, and aim to find a path to the goal state $[-1.35, 1.5, 0, 0]$. 

The Fig.~\ref{fig:bi_rs} depicts the details of node expansion and RSExpansion in HJBA*.
In forward search stage, $P(\mathbf{z}_0 \rightarrow \mathbf{z}^{\mathcal{S}}_i)$, the objective is to find a collision-free path connected to a  $\mathbf{z}_{\textup{best}}$ state such that there exists a $\mathcal{RS}(\mathbf{z}_{\textup{best}}, \mathbf{z}^{\mathcal{S}}_{i}, \mathbf{z}_{0})$.
By utilizing connected states $\mathbf{z}^{\mathcal{S}}_{i}$ picked from safe reachable set, HJBA* finds the $P(\mathbf{z}_0 \rightarrow \mathbf{z}^{\mathcal{S}}_i)$, demonstrated by the black and red segments in Fig.~\ref{fig:bi_rs_show}, in a fast manner while incurring minimal node expansions.
As mentioned previously, our node expansion strategy described in Fig.~\ref{alg:BA*_NodeExpand} and depicted in Fig.~\ref{fig:motion_premitives} entails the planning to search towards to $\mathbf{z}^{\mathcal{S}}_{i}$ initially.
If an obstacle is encountered, a backward node expansion is conducted to adjust the agent's pose, followed by a forward to the connected states $\mathbf{z}^{\mathcal{S}}_{i}$ again.
Analogously, in $P(\mathbf{z}^\mathcal{S}_i \leftarrow \mathbf{z}_g)$, the black segment is relatively short, the green segment, $|\mathcal{RS}(\mathbf{z}_{\textup{best}}, \mathbf{z}^{\mathcal{S}}_{i}, \mathbf{z}_{g})|$ constitutes a large proportion of the parking path $P(\mathbf{z}^\mathcal{S}_i \leftarrow \mathbf{z}_g)$, making the task less challenging.
As a result, there are only a few RSExpansion trials.
The findings align with our expectations.

For traditional motion primitives, when nodes expand forward, they also expand backward, resulting in some unnecessary expansions.
Autonomous parking is typically carried out in narrow environments, making it challenging for resolution-based search algorithms to locate $\mathbf{z}_{\textup{best}}$ in the region near the parking spot.
Therefore, finding a collision-free $\textup{RSExpansion}(\mathbf{z}_{\textup{best}}, \mathbf{z}_g)$ can take a long time or result in a timeout. 
Using a higher resolution map or tuning parameters would mitigate this problem but it is not solving the problem fundamentally.
HJBA* solve this challenge by utilizing the connected states in safe reachable set $\mathcal{S}$.
The connected states not only increase the proportion of $|\mathcal{RS}|$ but also make the search problem easier.

\subsubsection{Multiple parking requests demonstration}
In order to showcase the HJBA*'s capability of processing multiple parking requests, we provide two practical scenarios where HJBA* is validated. 
The two parking scenario involves a driver submits parking requests from various starting positions and HJBA* generates parking solutions in real-time.

\begin{figure}[!ht]\
  \centering
    \begin{subfigure}{\linewidth}
        \includegraphics[width=\linewidth]{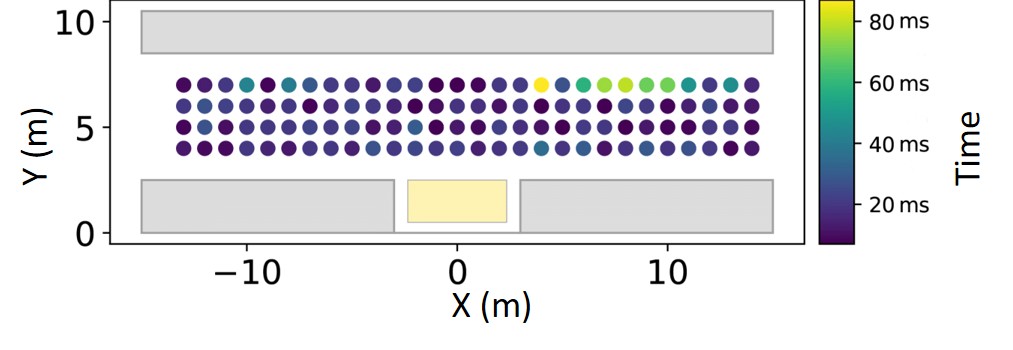}
    \label{fig:comprehensive}
  \end{subfigure}
      \begin{subfigure}{\linewidth}
        \includegraphics[width=\linewidth]{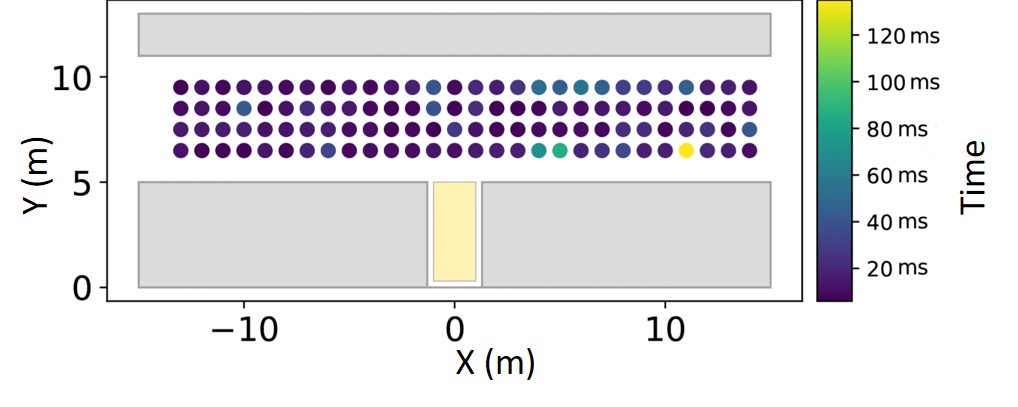}
    \label{fig:comprehensive_p}
  \end{subfigure}
  \caption{Solution time of HJBA* in parallel parking (top) and perpendicular parking (bottom) under different parking requests.}
  \label{fig:compreh}
\end{figure}

As illustrated in Fig.~\ref{fig:compreh}. The result reveals that HJBA* exhibits a reliable and effective capacity to manage multiple parking requests.
It should be noted that some solutions took longer compared to others.
This can be attributed to the uncertainty of connected states, whereby a car's proximity to the obstacle boundary makes it more challenging to find a direct path to the connected states as compared to other positions within the parking lot. Nevertheless, the computation time is still within an acceptable range, and the overall performance remains stable.

\subsubsection{Handling Narrow and Complex Parking Scenarios}
In this section, we demonstrate the versatility of our algorithm by showcasing its effectiveness not only in regular parking scenarios but also in challenging environments with clustered, complex, and narrow parking conditions.

\begin{figure}
\centering
\includegraphics[width=1\linewidth]{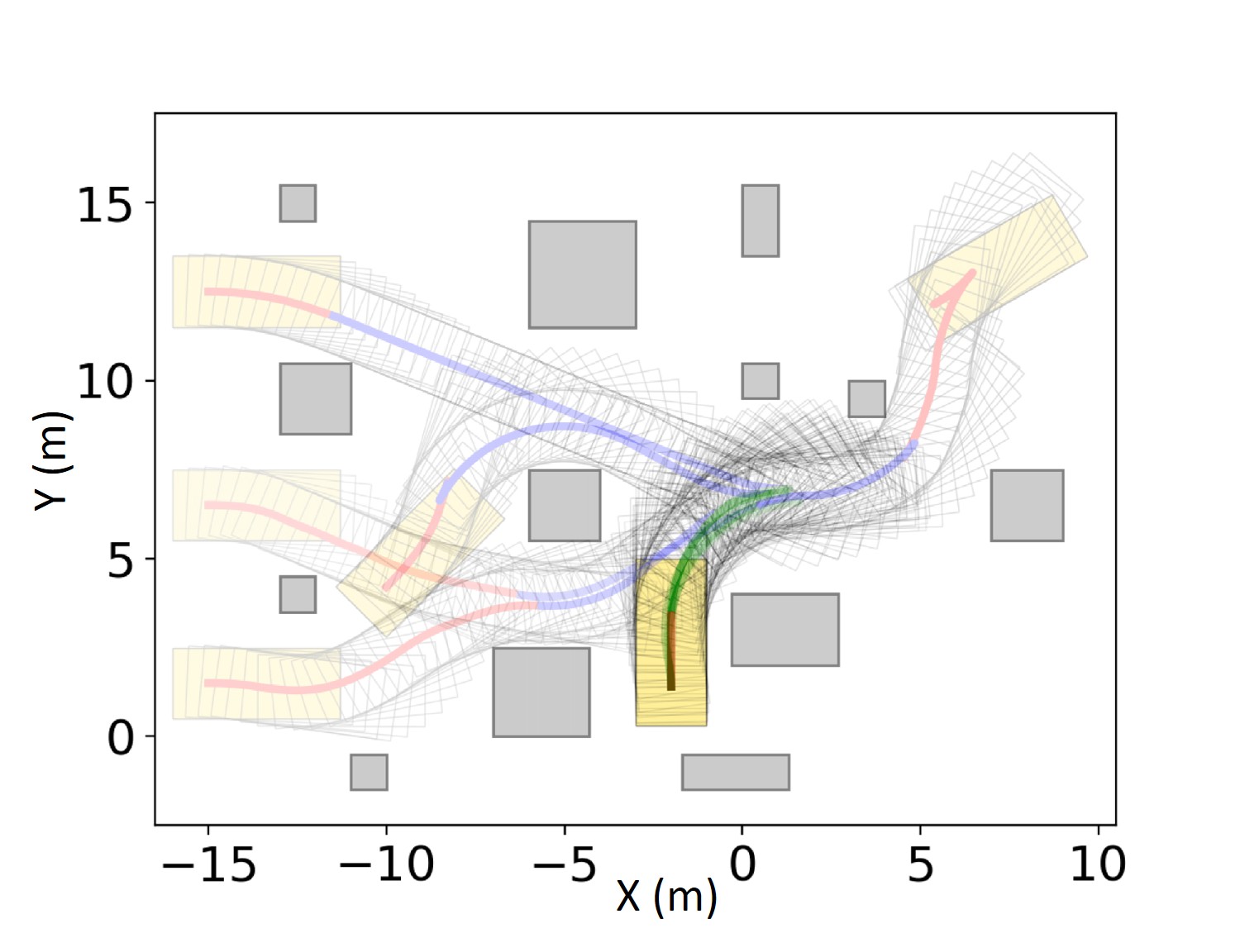}
\caption{The demonstration of complex parking scenarios}
\label{fig:irr}
\end{figure}

Figure~\ref{fig:irr} illustrates the setup we used for evaluation. The initial parking pose comprises five positions: $\mathbf{z}_0=[-15, 1.5, 0.0]$, $\mathbf{z}_0=[-15, 6.5, 0.0]$, $\mathbf{z}_0=[-15, 12.5, 0.0]$, $\mathbf{z}_0=[-10, 4.2, \pi/4]$, and $\mathbf{z}_0=[6.0, 12.5, \pi/6]$. The blue and green paths represent the $\mathcal{RS}$ curves generated by our algorithm, which are computationally efficient.

With HJBA* algorithm, we successfully identify the suboptimal pose $\mathbf{z}_{\textup{best}}$ that exhibits collision-free $\mathcal{RS}$ curves towards the connected points $\mathbf{z}^{\mathcal{S}}_k$. This highlights the capability of our approach to handle various parking scenarios, including those with obstacles clustered in narrow spaces.

A significant challenge faced by search-based algorithms in obstacle-clustered and narrow environments is the need to adjust parameters such as grid size or heuristic parameters. These adjustments often come at the cost of increased computation time or require substantial human effort.

However, it is important to note that for our large-scale testing, we specifically focus on classic parking scenarios to ensure a controlled evaluation environment.

\subsection{Performance Comparisons on Regular Parking Scenarios}

To thoroughly evaluate the effectiveness of our HJBA* algorithm across various regular parking scenarios and compare it with state-of-the-art parking planning algorithms, we have established four key evaluation criteria:
\begin{enumerate}[i]
	\item {Computation Time}: The computation time refers to the duration it takes for an algorithm to return a solution for a given parking scenario. Offline computation times, such as the heuristic function for HA* and MHHA*, as well as the BRT for HJBA*, are excluded.
	
	\item {Node Numbers}: The computation time is directly reflected by the number of node expansions, which is indicative of both the rationality of the heuristic function and the structure of the algorithm.
	
	\item {Path Length}: The path length is calculated as the summation of Euclidean distances between pairs of nodes along the solution.
	
    \item {Direction Changes}: When the parking environment is narrow, the parking maneuvers will be composed of forward and backward movements at the cusp.
	
	\item {Failure Rate}: We define a failure condition for the algorithm where it is deemed unsuccessful if the number of nodes in the closed list exceeds a threshold value, denoted by $N$. 
    The specific value of $N$ is determined by the complexity of the parking scenario.
\end{enumerate}

\subsubsection{Computation time}
\begin{figure*}[!ht]
  \centering
    \begin{subfigure}{0.32\linewidth}
        \includegraphics[width=\linewidth]{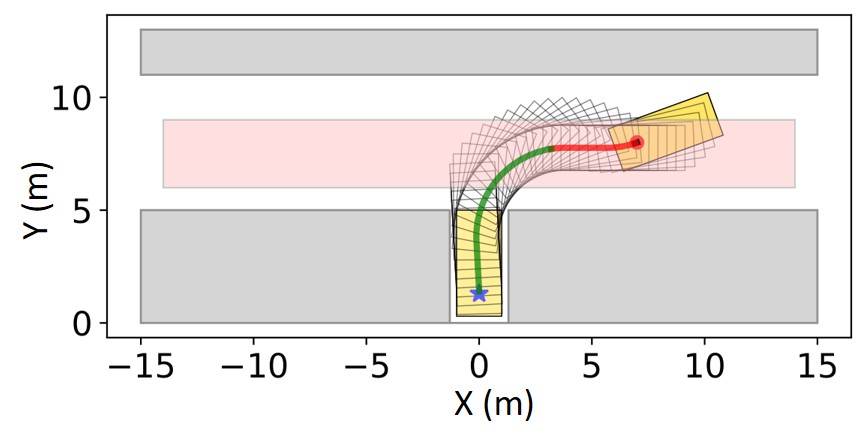}
        \caption{Free-space perpendicular parking}
    \label{fig:perpendicular_random_exampleS1}
  \end{subfigure}
  \begin{subfigure}{0.32\linewidth}
    \centering
        \includegraphics[width=\linewidth]{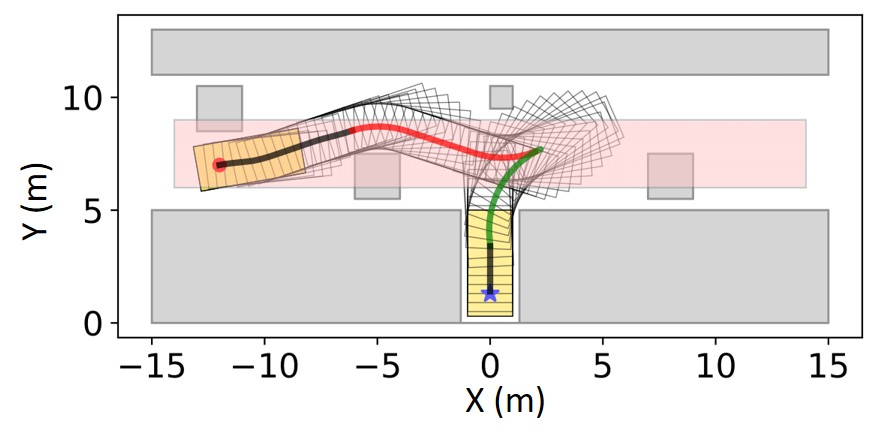}
        \caption{Constrained-space perpendicular parking}
    \label{fig:perpendicular_random_exampleS2}
  \end{subfigure}  
  \begin{subfigure}{0.32\linewidth}
        \includegraphics[width=\linewidth]{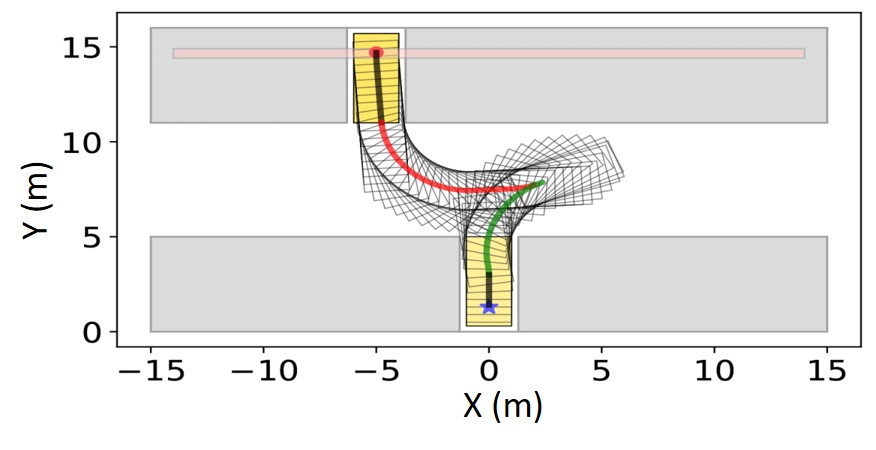}
        \caption{Position swapping in perpendicular parking}
    \label{fig:perpendicular_random_exampleS3}
  \end{subfigure}  
  \quad
  \begin{subfigure}{0.32\linewidth}
        \includegraphics[width=\linewidth]{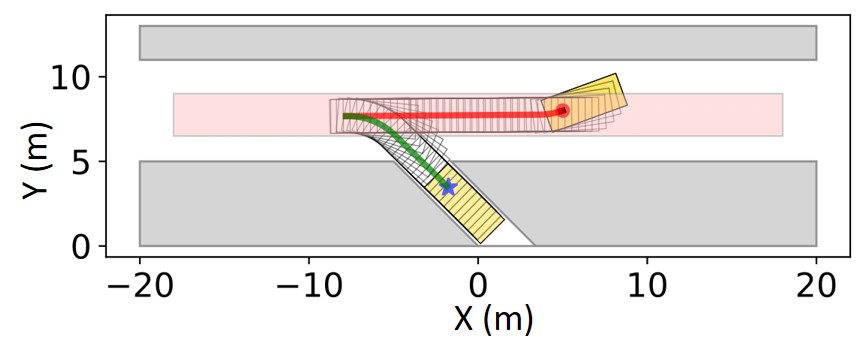}
        \caption{Free-space angle parking}
    \label{fig:angle_random_exampleS1}
  \end{subfigure} 
  \begin{subfigure}{0.32\linewidth}
        \includegraphics[width=\linewidth]{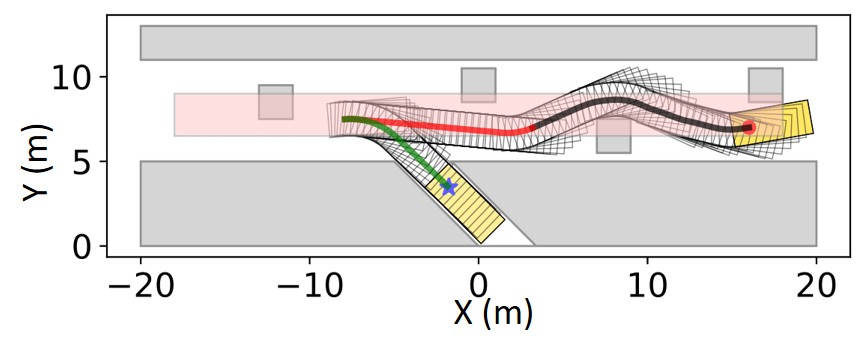}
        \caption{Constrained-space angle parking}
    \label{fig:angle_random_exampleS2}
  \end{subfigure} 
  \begin{subfigure}{0.32\linewidth}
        \includegraphics[width=\linewidth]{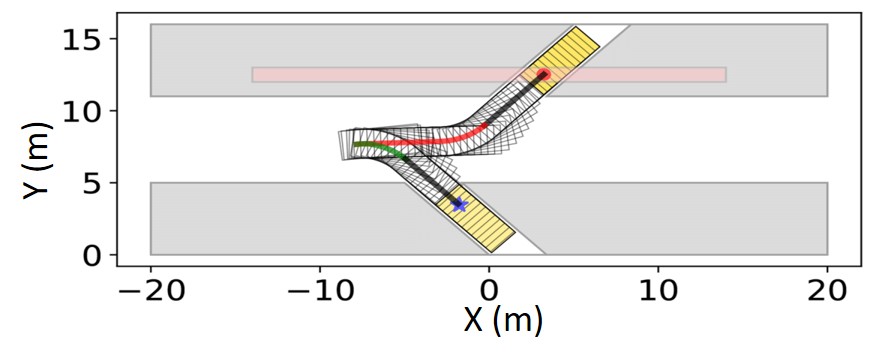}
        \caption{Position swapping in angle parking}
    \label{fig:angle_random_exampleS3}
  \end{subfigure} 
  \quad
    \begin{subfigure}{0.32\linewidth}
        \includegraphics[width=\linewidth]{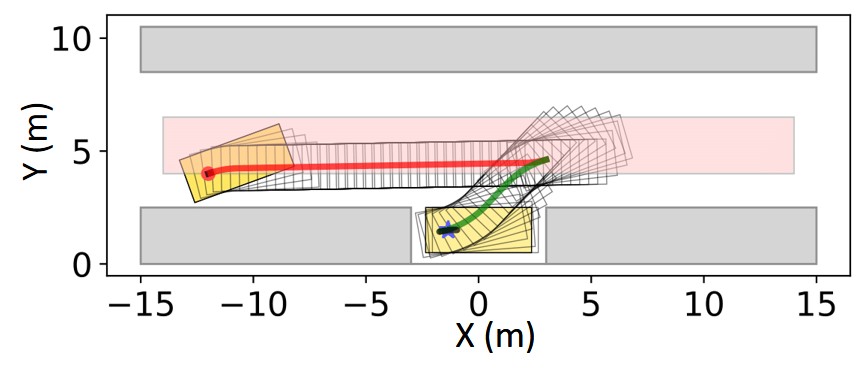}
        \caption{Free-space parallel parking}
    \label{fig:parallel_random_exampleS1}
  \end{subfigure} 
    \begin{subfigure}{0.32\linewidth}
        \includegraphics[width=\linewidth]{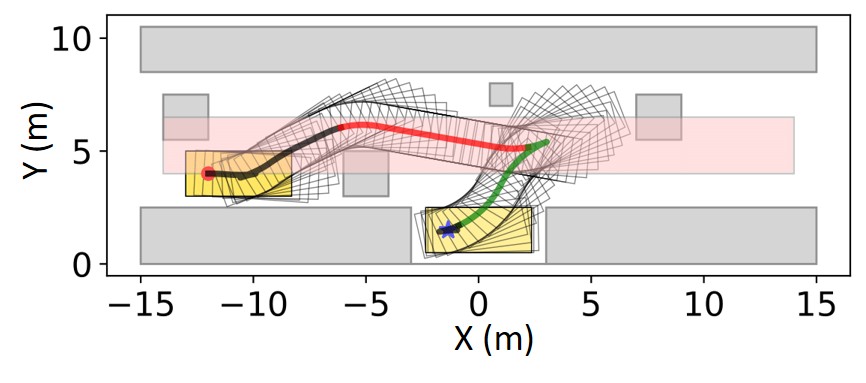}
        \caption{Constrained-space parallel parking}
    \label{fig:parallel_random_exampleS2}
  \end{subfigure} 
    \begin{subfigure}{0.32\linewidth}
    \includegraphics[width=\linewidth]{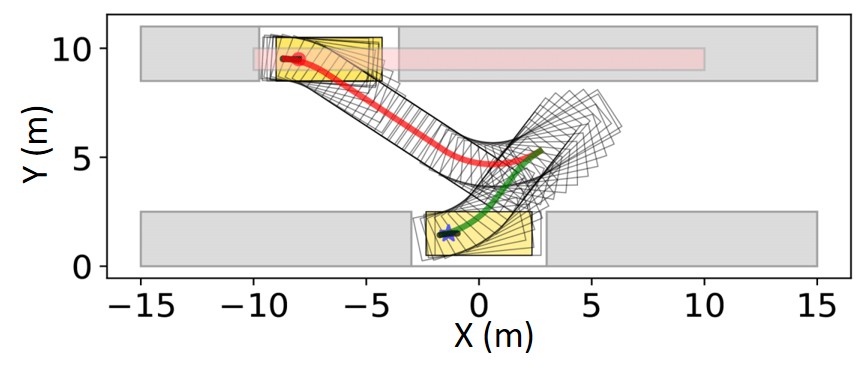}
        \caption{Position swapping in parallel parking}
    \label{fig:parallel_random_exampleS3}
  \end{subfigure} 
  
  \caption{The illustration of HJBA* in general parking scenarios.
  The red point is the initial pose and is randomly generated in the pink area.
  In common perpendicular parking, 
  The initial pose $x_{0}\in [-14, 14]$, $y_{0}\in [6.0, 9.0]$, $\theta_{0}\in [0, \pi/4]$.
  In common angle parking, we consider $x_{0}\in [-18, 18]$, $y_{0}\in [6, 8]$, $\theta_{0}\in [0, \pi/9]$.
  In parallel parking, the random initial parking pose region is $x_{0}\in [-14, 14]$, $y_{0}\in [4, 6.5]$, $\theta_{0}\in [0, \pi/9]$.}
  \label{fig:one case}
\end{figure*}  

The batch simulation results of 900 tests indicate that our HJBA* algorithm demonstrates a consistent computation speed in a general parking scenario, as presented in Tab.~\ref{tab:all_trials}.
A typical parking path $P(\mathbf{z}_{0}, \mathbf{z}_{g})$ comprises of two segments, namely $\mathcal{A}(\mathbf{z}_{0}, \mathbf{z}_{\text{best}})$ and $\mathcal{RS}(\mathbf{z}_{\text{best}}, \mathbf{z}_{g})$, with most of the computational effort dedicated to the former due to node expansion in space.
While HA* and MHHA* can sometimes provide a solution quickly, as evidenced by their minimal computation times, the real challenge in solving a parking problem lies in providing quick solutions for different initial states.
Our HJBA* algorithm maintains the fastest average computation speed across all scenarios.
In the case of parallel parking, the slow computation speed of HA* and MHHA* is attributed to the excessive node expansion required to find the optimal $\mathbf{z}_{\text{best}}$.
Even after optimizing the heuristic functions, they still need to expand a large number of nodes to reach the proximity of the parking spots.
As they approach the vicinity of a parking spot, the failure rate increases and the search time is extended since the heuristic cost for each node around this area is similar, and without guidance, HA* and MHHA* have to expand more nodes to locate the optimal $\mathbf{z}_{\text{best}}$.

While for our algorithm HJBA*, we have such connected points which can help to solve a parking problem in a fast manner by the bidirectional mechanism.
In our HJBA*, $P^{\textup{BA*}}_{(\mathbf{z}_{0}, \mathbf{z}_{g})} = P(\mathbf{z}_0 \rightarrow \mathbf{z}^\mathcal{S}_i) \oplus P(\mathbf{z}^\mathcal{S}_i \leftarrow \mathbf{z}_g)$.
As demonstrated in Fig.~\ref{fig:perpendicular_random_exampleS2}, the target of \emph{forward search} is to find $\mathbf{z}^{\textup{F}}_{\text{best}}$ and connect $\mathbf{z}^{\textup{F}}_{\text{best}}$ to $\mathbf{z}^\mathcal{S}_i$ with collision-free RS curves.
The red segment is $\mathcal{RS}(z^{\text{F}}_{\text{best}}, \mathbf{z}^\mathcal{S}_i)$, which saves a lot of computation time for the parking problem.
For the worst case, HJBA* has a smaller computation time compared to HA* and MHHA*.
In worse cases of obstacles clustered parking scenarios, the computation time is larger mainly due to the increase of $\left|\mathcal{A}(\mathbf{z}_{0}, \mathbf{z}^\mathcal{S}_i) \right|$.

\subsubsection{Node Numbers}
Still from Tab.~\ref{tab:all_trials}, our algorithm HJBA* has the minimum mean node numbers for all parking scenarios.
The trend is similar to the computation time.
In parallel parking, the performance of HA* and MHHA* become bad because they are not stable for any random initial states, in which our HJBA* is not affected with the help of $\mathbf{z}^\mathcal{S}_i$.
Nevertheless, in some simple random cases, HA* and MHHA* can provide a solution with one node while HJBA* needs at least two nodes due to the introduced sampling states.

\subsubsection{Path Length}
HA*, MHHA* and our HJBA* are classified as sub-optimal search algorithms since the path of $\mathcal{A}(\mathbf{z}_{0}, \mathbf{z}_{\text{best}})$ segment guided by the "cost-to-go" and "cost-to-come" is not guaranteed to be the optimal expansion.
However, the segment of $\mathcal{RS}(\mathbf{z}_{\text{best}}, \mathbf{z}_{g})$ is analytical optimal.
Theoretically, for a collision-free path $P$, a higher proportion of $\mathcal{RS}$ curves corresponds to a shorter path length.
In both perpendicular and parallel parking scenarios, our HJBA* algorithm achieves the shortest mean path length while maintaining the fastest computation speed. 
However, in angle parking, the mean path length of HJBA* is longer. This is because of the introduced connected states and bidirectional search mechanism, resulting in two $\mathcal{RS}$ curves, namely $\mathcal{RS}(\mathbf{z}^{\text{F}}_{\text{best}}, \mathbf{z}^\mathcal{S}_i)$ and $\mathcal{RS}(\mathbf{z}^{\text{B}}_{\text{best}}, \mathbf{z}^\mathcal{S}_i)$, in the solution of HJBA*.
The sum of these two curves can sometimes be greater than a direct $\mathcal{RS}$ curve. In this regard, the HJBA* algorithm balances optimality and computation time, which is also evident in the case of angle parking.
\begin{table*}[ht!]
    \centering

\caption{Evaluation indexes on different parking scenarios and variants trials$^{\text{2}}$.}
\label{tab:all_trials}
\begin{threeparttable}
\resizebox{\textwidth}{!}{\revised{
\begin{tabular}{c c|c|c c c|c c c|c c c|c c c}
\hline
\multirow{2}{*}{Scenarios$^{\text{1}}$} & \multirow{2}{*}{Algorithms}  & Failure  & \multicolumn{3}{c|}{Computation Time ($ms$)} & \multicolumn{3}{c|}{Node Numbers} & \multicolumn{3}{c|}{Path Length ($m$)}  & \multicolumn{3}{c}{\revised{Direction Changes}}\\ \cline{4-15}
                                        &             &   Rate      &  mean    & min    &max                                   & mean &min &max                   &mean  &min &max &\revised{mean} &\revised{min} &\revised{max}\\ 
\hline
\multirow{3}{*}{$\mathbf{a}$} & HA*                  & $0\%$     &568   &106  &1520                                  &372 &90  &870               &26.9 &25.3  &28.3      &1&0&1           \\ 
& MHHA*                                               & $0\%$     &285   &133  &573                                    &213 &105  &394                &26.7 &25.1  &28.3   &1&0&1             \\ 
& Ours                                                & $0\%$     &\textbf{27} &\textbf{10}  &\textbf{52}   &\textbf{7}  &\textbf{2}  &\textbf{21}   &\textbf{23.9}&\textbf{21.7}&\textbf{26.7} &\textbf{1}&\textbf{0}&\textbf{1} \\ 
\hline
\multirow{3}{*}{$\mathbf{b}$} & HA*                  & $0\%$     &947 &5  &3973                    &422 &\textbf{1}  &1598               &12.5 &6.6 &22.5     &\textbf{1}&1&3         \\ 
& MHHA*                                               & $0\%$     &732 &\textbf{4}  &2430                             &349 &\textbf{1}  &1024                 &12.4 &\textbf{6.6}  &23.4  &1&1&3 \\ 
& Ours                                               & $0\%$     &\textbf{63} &9  &\textbf{306}                 &\textbf{23} &2  &\textbf{80}     &\textbf{12.0} &8.1  &\textbf{20.8} &2&\textbf{1}&\textbf{3} \\ 
\hline
\multirow{3}{*}{$\mathbf{c}$} & HA*                  & $0\%$     &1135 &393  &1952                                 &399   &149  &510            &23.7&\textbf{17.9}  &30.6      &1&1&1       \\ 
& MHHA*                                               & $0\%$     &1058 &24  &2253                                     &425 &12   &1002           &24.5 &\textbf{17.9}  &30.6      &1&1&1      \\ 
& Ours                                               & $0\%$     &\textbf{85}  &\textbf{23}  &\textbf{144}  &\textbf{25}  &\textbf{7}  &\textbf{40}      &\textbf{22.8} &18.0  &\textbf{29.4}  &\textbf{1}&\textbf{1}&\textbf{1}    \\ 
\hline
\hline
\multirow{3}{*}{$\mathbf{d}$} & HA*                  & $5\%$       &852   &10  &3727           &512 & 7 &1643         &15.0 &\textbf{8.1}  &29.1    &\textbf{1}&0&\textbf{1}                     \\ 
& MHHA*                                               & $14\%$     &1335   &9  &4085             &688 &7  &1574         &\textbf{13.7} &\textbf{8.1}  &\textbf{28.1}  &1&0&2     \\ 
& Ours                                 & \bm{${0\%}$}     &\textbf{82} &\textbf{11} &\textbf{280}    &\textbf{13} &\textbf{2} &\textbf{79}    &17.2 & 9.3 & 34.0    &3&\textbf{1}&3      \\ 
\hline
\multirow{3}{*}{$\mathbf{e}$} & HA*               & $2\%$     &901 &8  &8186             &454 &\textbf{1}  &1521          &17.4 &\textbf{8.2}  &\textbf{34.2}  &\textbf{1}&\textbf{0}&\textbf{3}     \\ 
& MHHA*                                               & $1\%$     &1214 &\textbf{6}  &6955             &523&\textbf{1}  &1463          &\textbf{16.4} &\textbf{8.2}  &38.1 &1&0&3 \\ 
& Ours                                               & \bm{${0\%}$}      &\textbf{115} &9  &\textbf{885}       &\textbf{32} &2  &\textbf{176}           &19.2 &9.1  &37.3       &4&2&7    \\ 
\hline
\multirow{3}{*}{$\mathbf{f}$} & HA*                  & $0\%$     &534 &46  &2239                                    &169   &13  &1169              &21.8 &\textbf{15.7}  &31.5      &\textbf{1}&1&\textbf{1}        \\ 
& MHHA*                                               & $0\%$     &387 &44  &3642                                     &117   &13   &1287             &\textbf{21.0} &\textbf{15.7}  &\textbf{27.0} &1&1&1 \\ 
& Ours                                               & \bm{${0\%}$}      &\textbf{187}  &\textbf{22}  &\textbf{429}       &\textbf{21}  &\textbf{2}    &\textbf{55}           &22.0 &15.9  &30.2     &4&\textbf{1}&4      \\ 
\hline
\hline
\multirow{3}{*}{$\mathbf{g}$} & HA*                  & $58\%$     &1826   &96  &5632                                   &940 &89  &1644               &16.0 &\textbf{7.7}  &26.0     &\textbf{3}&\textbf{2}&\textbf{4}         \\ 
& MHHA*                                               & $35\%$     &2622   &237  &7613                                 &1156 &163  &2524               &17.0 &\textbf{7.7}  &26.8      &3&2&5       \\ 
& Ours                                               & \bm{${0\%}$}      &\textbf{48} &\textbf{11}  &\textbf{122}              &\textbf{17}  &\textbf{8}  &\textbf{24}         &\textbf{14.4}&7.8 &\textbf{22.0} &3&2&4   \\ 
\hline
\multirow{3}{*}{$\mathbf{h}$} & HA*                  & $34\%$     &2484 &286 &7201           &875 &141  &2017                 &14.7 &7.2  &32.2      &4&2&5        \\ 
& MHHA*                                               & $37\%$     &1552 &249  &4739                                     &613 &140  &1455               &14.0 &\textbf{6.9}  &28.7       &4&2&5       \\ 
& Ours                                              & \bm{${0\%}$}      &\textbf{81} &\textbf{13}  &\textbf{278}    &\textbf{21} &\textbf{8}  &\textbf{84}  &\textbf{14.0} &7.5  &\textbf{25.7}   &\textbf{3}&\textbf{2}&\textbf{5} \\ 
\hline
\multirow{3}{*}{$\mathbf{i}$} & HA*                  & $49\%$     &5581 &142  &10793                                    &2163   &87  &3929            &24.2 &18.0  &34.7      &6&4&6         \\ 
& MHHA*                                               & $23\%$     &3298 &272  &8139                                    &1531 &196   &3213             &23.2 &18.9 &29.9          &6&4&\textbf{6}      \\ 
& Ours                                 & \bm{${0\%}$}      &\textbf{199}  &\textbf{23}  &\textbf{564}   &\textbf{33}  &\textbf{17}    &\textbf{114}  &\textbf{20.1} &\textbf{17.5}  &\textbf{25.3} &\textbf{5}&\textbf{4}&8   \\ 
\hline
\end{tabular}}}
\end{threeparttable}
\begin{tablenotes}
\footnotesize
\item{1} The letters in scenarios correspond to the letters in Fig. \ref{fig:one case}.
\item{2} The evaluation indexes of failure cases are not included. 
\end{tablenotes}
\end{table*}

\revised{\subsubsection{Direction Changes}
The number of cusps in a parking path is typically associated with the path length rather than the computation time. Our HJBA* algorithm demonstrates a comparable number of cusps to HA* and MHHA* in perpendicular and parallel parking scenarios, resulting in a superior mean path length for HJBA*. However, in angle parking situations, HJBA* exhibits a higher number of cusps compared to HA* and MHHA*.

This discrepancy can be attributed to the relationship between the path length and the number of cusps. In angle parking, the involvement of two $\mathcal{RS}$ curves to construct the parking path leads to an increase in the number of cusps. Conversely, the similar number of cusps observed in perpendicular and parallel parking can be attributed to their reverse parking nature. It is important to note that angle parking in our study exclusively involves forward parking, which contributes to the higher number of cusps observed in HJBA* for this scenario.}

\subsubsection{Failure rate}
The $0 \%$ failure rate achieved by our HJBA* algorithm in collision-free perpendicular parking scenarios demonstrates its capability to find solutions with the same effectiveness as HA* and MHHA*. However, the failure rates for HA* and MHHA* in angle and parallel parking are comparatively high. While parameter tuning of grid size, step size, and heuristic function may improve their success rates, it is a time-consuming process, and parameter tuning alone cannot guarantee a stable planner. Notably, the failure rate of HA* in random common parallel parking is even higher than that in clustered parallel parking scenarios, as the obstacles in the latter make some initial parking states closer to the parking goal pose, thereby improving the success rate of HA*. In contrast, our HJBA* algorithm exhibits extremely stable performance with a $0.0\%$ failure rate across all tests.

\subsection{Performance Comparisons on Irregular Parking Scenarios}
\begin{figure}[t]
    \centering
    \includegraphics[width=\linewidth]{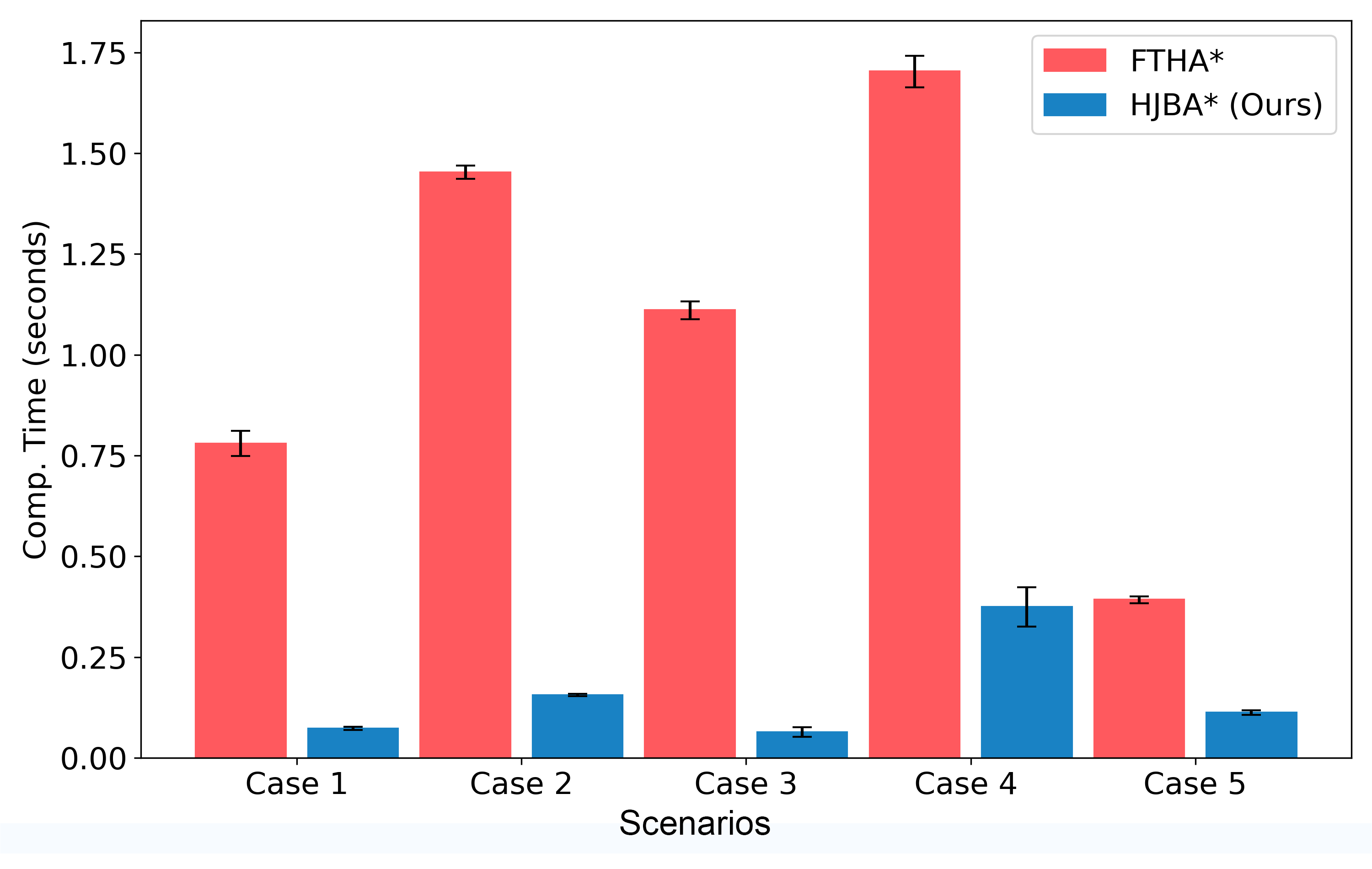}
    \caption{Comparison of algorithms FTHA*~\cite{bai2022irregularly} and HJBA* across random scenarios.
    Each case is repeatedly run five times.
    }
    \label{fig:comparison_vs_ftha}
\end{figure}
To further evaluate our algorithm, we compare it with state-of-the-art parking path planning algorithms~\cite{bai2022irregularly} on irregular parking scenarios.
The path planning layer typically serves as a sub-module in trajectory optimization.
An optimization-based trajectory planner generally includes a path planning layer that provides an initial guess for lower-layer nonlinear programming problems.
For this paper, we focus solely on the path planning module as a baseline, as the lower layer of trajectory optimization falls outside the scope of this paper.

\begin{figure*}
    \centering
    \begin{subfigure}[t]{\linewidth}
        \centering
        \includegraphics[width=\linewidth]{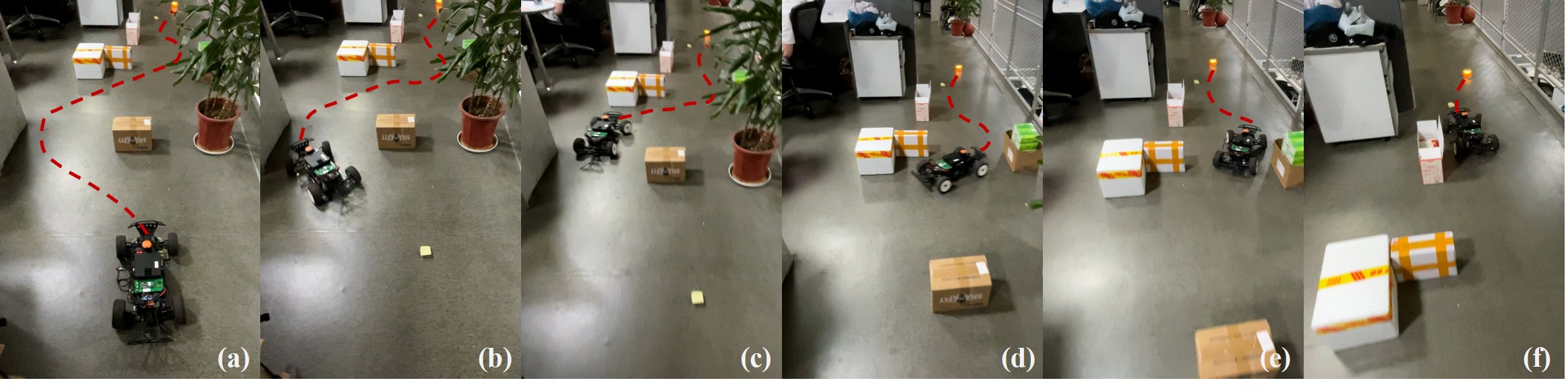}
    \end{subfigure}%
    \caption{HJBA*: Irregular parking snapshots in an obstacle-cluttered environment.
    The red trajectory is planned by HJBA* and tracked by the 1:10 scale autonomous platform.
    The car avoids polytopic obstacles and arrives at the goal position.
    }
    \label{fig:exp_car_traj}
\end{figure*}
We randomly generate five irregular parking scenarios\footnote{The randomly generated irregular parking scenarios are based on the MATLAB code from the~\hyperlink{https://github.com/libai1943/ParkingMotionPlanningTITS21}{Github Link}.} and configure the vehicle and polytopic obstacles according to the parameters outlined in~\cite{bai2022irregularly}. 
Each case is repeatedly run five times.
The comparison results, shown in Fig.~\ref{fig:comparison_vs_ftha}, demonstrate the efficiency of our algorithm even in irregular parking scenarios. Replacing the path planning layer with our HJBA* improves trajectory optimization efficiency by \(84.7\%\).

In comparison to another recent state-of-the-art parking path planning algorithm, Mirror the Parking Target (MPT) method~\cite{Jia2024mirror}, which takes 74 milliseconds for parking scenarios without obstacle consideration, our HJBA* algorithm requires only 52 milliseconds, maintaining its competitiveness.
Interestingly, MPT also accounts for target reachability during its design.

\subsection{Real World Experiments}
\begin{figure}
    \centering
    \begin{subfigure}[t]{\linewidth}
        \centering
        \includegraphics[width=\linewidth]{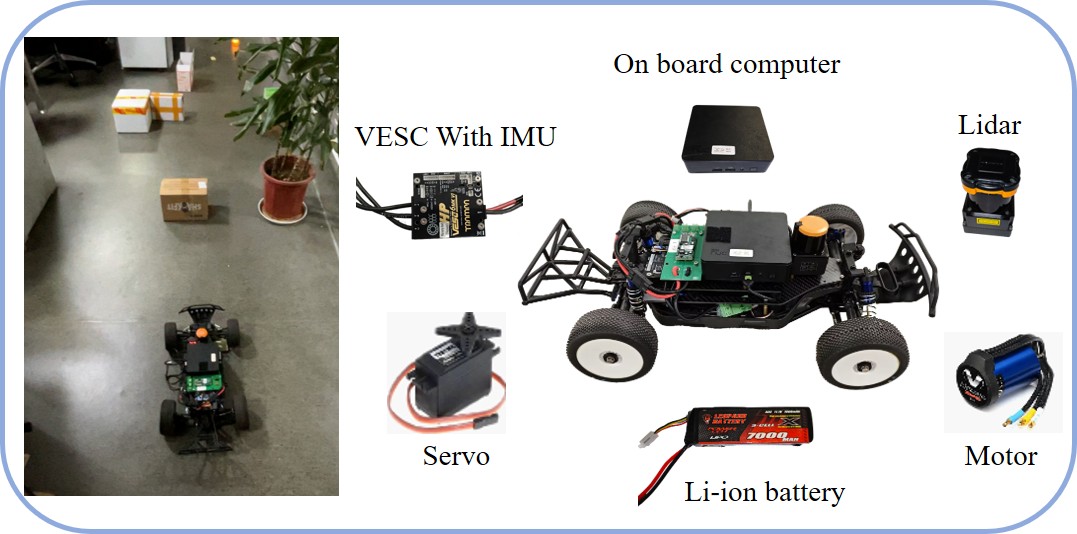}
    \end{subfigure}%
    \caption{The irregular parking map and 1:10 scale autonomous platform.}
    \label{fig:exp_car_fig}
\end{figure}

Real-world experiments have been conducted to further validate the effectiveness of the proposed algorithm in irregular parking scenarios. The vehicle platform and experimental setup are shown in Fig.~\ref{fig:exp_car_fig}. The sensor suite includes a Hokuyo UST-10LX LiDAR with a maximum scan frequency of 40~Hz. The LiDAR measurements provide discrete points that represent the surrounding environment and obstacles in the parking scene. Odometry is provided by the Vedder Electronic Speed Controller (VESC). Vehicle localization and state estimation are obtained through Cartographer by fusing LiDAR measurements, the IMU embedded in the VESC, and odometry. In the control loop, a pure-pursuit controller is implemented to track the reference trajectory generated by the planner based on the estimated vehicle state. The results shown in Fig.~\ref{fig:exp_car_traj} demonstrate the effectiveness of the proposed method: the vehicle successfully avoids all polytopic obstacles and reaches the target parking position. Additional experimental details can be found in the accompanying video\footnote{\url{https://youtu.be/lafie7BW9oE}.}.

\section{CONCLUSIONS}\label{sec: conclusion}
In this paper, we proposed HJBA*, a two-layer planning framework for fast and reliable autonomous parking in tight and cluttered environments. The proposed method combines offline Hamilton--Jacobi reachability analysis with online bidirectional A* search, where a safe reachable set is constructed by intersecting the backward reachable tube with the collision-free safe set, and sampled connected states from this set are used to guide the online search process. In this way, HJBA* explicitly incorporates reachability, obstacle-aware safety, and vehicle kinematic feasibility into the planner, while effectively reducing ineffective node expansions in narrow parking regions. Extensive simulations and real-world experiments demonstrate that the proposed framework achieves a favorable balance among efficiency, robustness, and solution quality across a wide range of parking scenarios. In particular, HJBA* maintained a 0\% failure rate in 900 tests covering perpendicular, angle, and parallel parking, significantly improved trajectory-optimization efficiency in irregular parking scenarios, and was successfully validated on a real vehicle platform in obstacle-cluttered environments. These results indicate that integrating offline reachable guidance with online search is a practical and effective strategy for improving both planning reliability and real-time performance in autonomous parking. Nevertheless, the current work mainly focuses on static parking environments and the path-planning layer. Extending the proposed framework to dynamic obstacles, interactive multi-agent settings, and tighter integration with downstream optimization and control remains an important direction for future research.

\bibliographystyle{IEEEtran}
\bibliography{main.bib}


\begin{IEEEbiography}
[{\includegraphics[width=1in,height=1.25in,clip,keepaspectratio]{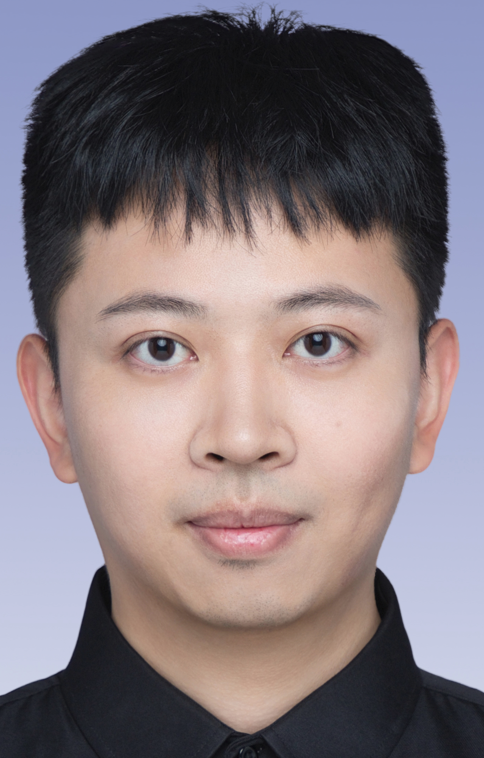}}]{Xuemin Chi}
received the B.Eng. degree in vehicle engineering in 2017 from Shenyang University of Technology, Shenyang, China, and the M.Sc. degree in vehicle engineering in 2019 from Dalian University of Technology, Dalian, China. 

He is currently working toward the Ph.D. degree in control engineering at Zhejiang University, Hangzhou, China.
His research interests include motion planning,
safe model predictive control algorithms.
\end{IEEEbiography}

\begin{IEEEbiography}
[{\includegraphics[width=1in,height=1.25in,clip,keepaspectratio]{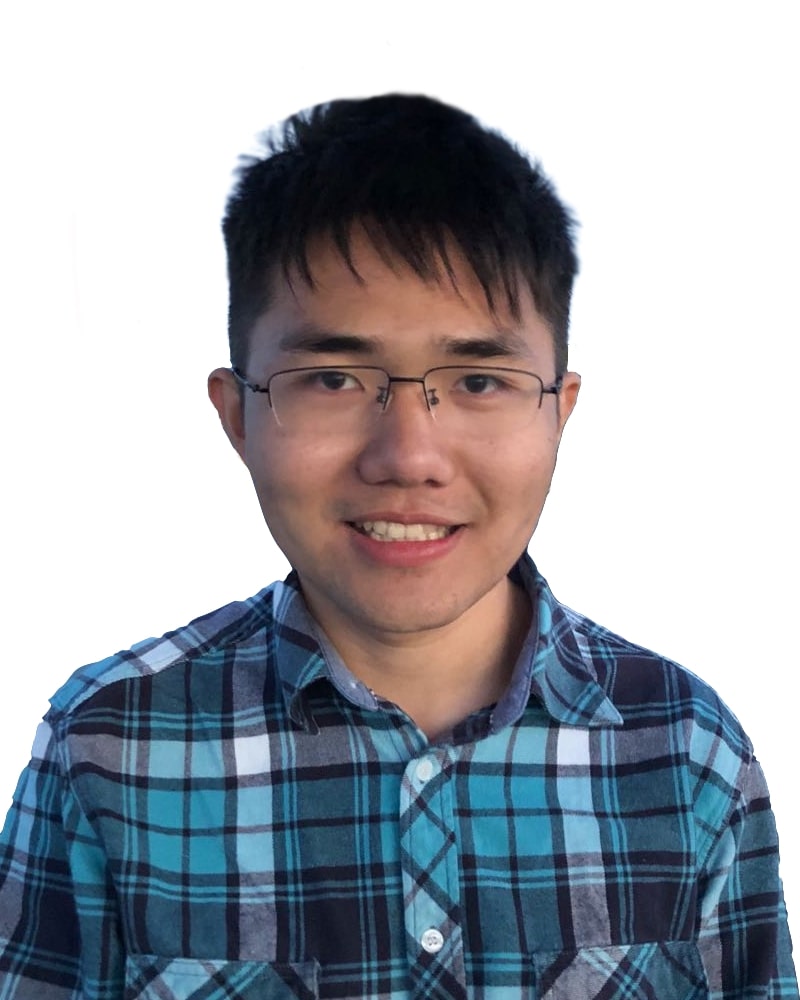}}]{Jun Zeng}
received his Ph.D. in Control and Robotics at the Department of Mechanical Engineering at University of California, Berkeley, USA in 2022 and Dipl. Ing. from Ecole Polytechnique, France in 2017, and a B.S.E degree from Shanghai Jiao Tong University (SJTU), China in 2016.
His research interests lie at the intersection of optimization, control, planning, and learning with applications on various robotics platforms.
\end{IEEEbiography}

\begin{IEEEbiography}
[{\includegraphics[width=1in,height=1.25in,clip,keepaspectratio]{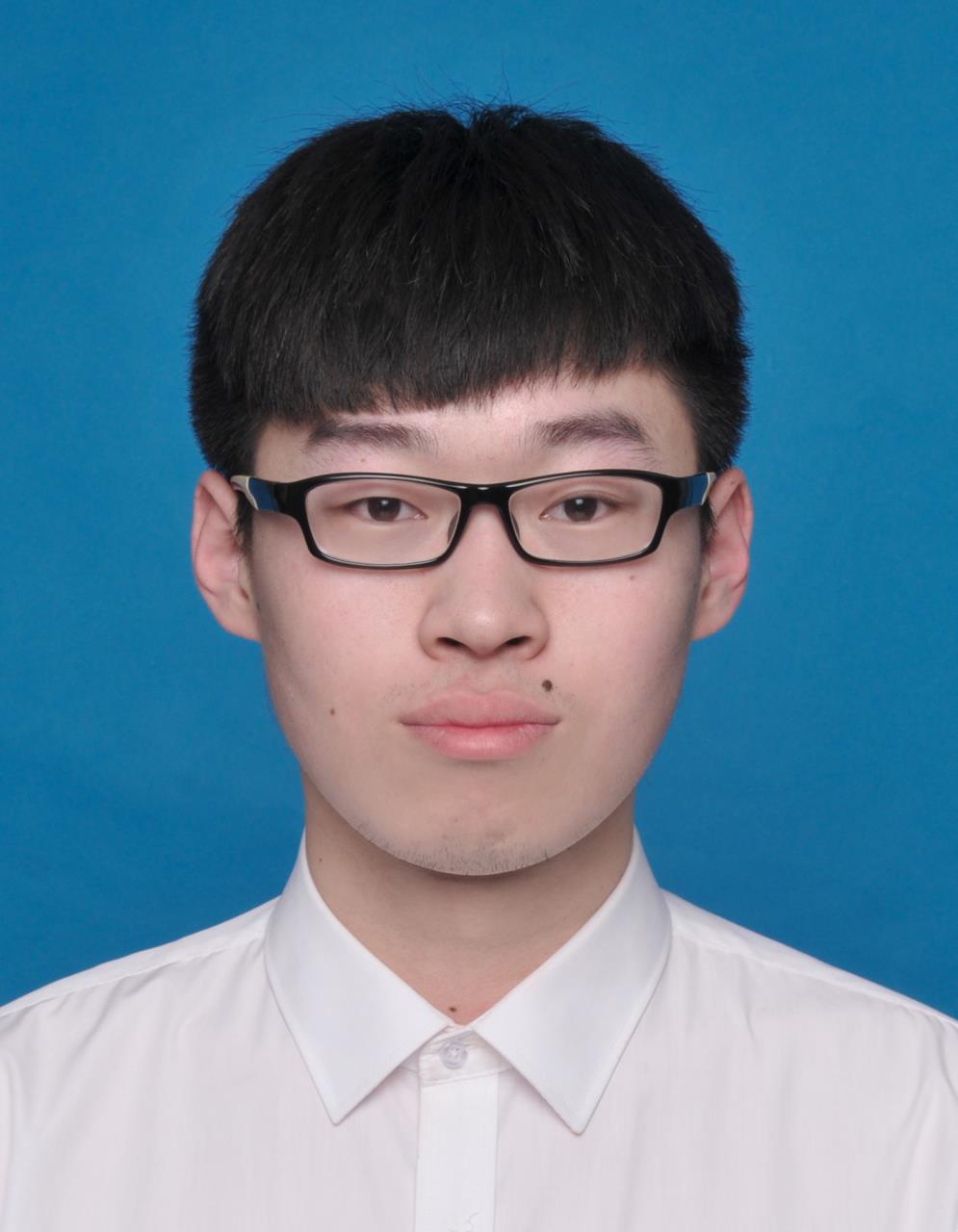}}]{Jihao Huang}
 received the B.Eng. degree in automation in 2020 from Hangzhou Dianzi University, Hangzhou, China. 
 
 He is currently pursuing the Ph.D. degree in control science and engineering at Zhejiang University, Hangzhou, China.His current research interests include multi-robot system, motion planning, and control theory.
\end{IEEEbiography}

\begin{IEEEbiography}
[{\includegraphics[width=1in,height=1.25in,clip,keepaspectratio]{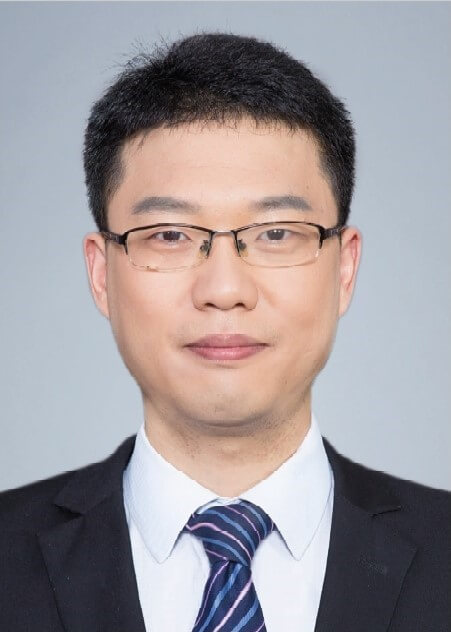}}]{Zhitao Liu}
 (M’13) received the B.S. degree from Shandong University, China, in 2005, and the Ph.D. degree in control science and engineering from Zhejiang University, Hangzhou, China, in 2010. 
 
 From 2011 to 2014, he was a Research Fellow with TUM CREATE, Singapore. He was an Assistant Professor from 2015 to 2016 and an Associate Professor from 2017 to 2021 in Zhejiang University, where he is currently a Professor with the Institute of Cyber-Systems and Control, Zhejiang University. His current research interest include robust adaptive control, wireless transfer systems and energy management systems.
\end{IEEEbiography}

\begin{IEEEbiography}
[{\includegraphics[width=1in,height=1.25in,clip,keepaspectratio]{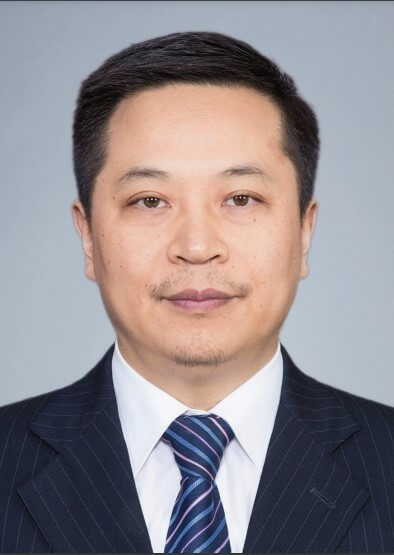}}]{Hongye Su}
 (SM’14) was born in 1969. He received the B.S. degree in industrial automation from the Nanjing University of Chemical Technology, Jiangsu, China, in 1990, and the M.S. and Ph.D. degrees in industrial automation from Zhejiang University, Hangzhou, China, in 1993 and 1995, respectively.

 From 1995 to 1997, he was a Lecturer with the Department of Chemical Engineering, Zhejiang University. From 1998 to 2000, he was an Associate Professor with the Institute of Advanced Process Control, Zhejiang University, where he is currently a Professor with the Institute of Cyber-Systems and Control. His current research interests include robust control, time-delay systems, and advanced process control theory and applications.
\end{IEEEbiography}

\end{document}